\documentclass[lettersize,journal]{IEEEtran}
\usepackage{subfiles}    

\usepackage{amsmath,amsfonts}
\usepackage{algorithmic}
\usepackage{algorithm}
\usepackage{array}
\usepackage[caption=false,font=normalsize,labelfont=sf,textfont=sf]{subfig}
\usepackage{textcomp}
\usepackage{stfloats}
\usepackage{url}
\usepackage{verbatim}
\usepackage{graphicx}
\usepackage{cite}

\usepackage{hyperref}
\usepackage{cleveref} 
\usepackage{latexsym}
\usepackage{makecell}
\usepackage{multirow}
\usepackage{array}
\usepackage{color}

\newcolumntype{L}[1]{>{\raggedright\arraybackslash}p{#1}}
\newcolumntype{C}[1]{>{\centering\arraybackslash}p{#1}}
\newcolumntype{R}[1]{>{\raggedleft\arraybackslash}p{#1}}

\begin{document}

\title{Learning from Rendering: Realistic and Controllable Extreme Rainy Image Synthesis for Autonomous Driving Simulation}

\author{Kaibin Zhou, Kaifeng Huang,~\IEEEmembership{Member,~ACM}, Hao Deng,~\IEEEmembership{Member,~IEEE}, Zelin Tao,\\ Ziniu Liu, Lin Zhang,~\IEEEmembership{Senior Member,~IEEE}, and Shengjie Zhao,~\IEEEmembership{Senior Member,~IEEE}
\thanks{

Kaibin Zhou, Kaifeng Huang, Hao Deng, Zelin Tao, Ziniu Liu, Lin Zhang and Shengjie Zhao are with the School of Computer Science and Technology, Tongji University, Shanghai 201804, China (email: kb824999404@163.com, \{kaifengh, denghao1984, 2311447, 2433289, cslinzhang, shengjiezhao\}@tongji.edu.cn).
}
}




\maketitle
\begin{abstract}

Autonomous driving simulators provide an effective and low-cost alternative for evaluating or enhancing visual perception models. However, the reliability of evaluation depends on the diversity and realism of the generated scenes. Extreme weather conditions, particularly extreme rainfalls, are rare and costly to capture in real-world settings. While simulated environments can help address this limitation, existing rainy image synthesizers often suffer from poor controllability over illumination and limited realism, which significantly undermines the effectiveness of the model evaluation.
To that end, we propose a learning-from-rendering rainy image synthesizer, which combines the benefits of the realism of rendering-based methods and the controllability of learning-based methods. To validate the effectiveness of our extreme rainy image synthesizer on semantic segmentation task, we require a continuous set of well-labeled extreme rainy images. By integrating the proposed synthesizer with the CARLA driving simulator, we develop CARLARain—an extreme rainy street scene simulator which can obtain paired rainy-clean images and labels under complex illumination conditions. Qualitative and quantitative experiments validate that CARLARain can effectively improve the accuracy of semantic segmentation models in extreme rainy scenes, with the models’ accuracy (mIoU) improved by $5\% - 8\%$ on the synthetic dataset and significantly enhanced in real extreme rainy scenarios under complex illuminations. Our source code and datasets are available at https://github.com/kb824999404/CARLARain/.

\end{abstract}

\begin{IEEEkeywords}
Rainy image synthesis, autonomous driving, simulation, visual perception, image deraining
\end{IEEEkeywords}

\section{Introduction}

\IEEEPARstart{I}{n} recent years, visual perception models \cite{chen2022vision,wang2023internimage,jain2023semask,hassani2023neighborhood} have advanced significantly in autonomous driving. The performance of visual perception models for autonomous driving systems relies on the diversity of datasets. However, due to the rare occurrence of extreme weather \cite{sakaridis2021acdc,sun2022shift}, the real-world efficacy of visual perception models is significantly impaired under such conditions. Extreme rainfalls \cite{westra2014future} (rain intensity $\geq 50mm/h$) is the most frequently occurring of extreme weather, which seriously interferes with the acquisition of scene environment information by visual perception models, thereby undermining the accuracy of tasks such as semantic segmentation, instance segmentation, etc.

To enhance the accuracy of visual perception models, it is crucial to include a sufficient number of extreme rainy images. However, acquiring labeled real extreme rainy images is challenging due to their high costs of collection and annotation, as well as the rarity of extreme rainfalls. Therefore, a more efficient alternative is to synthesize rainy images \cite{garg2006photorealistic,halder2019physics,wang2021rain} along with corresponding labels \cite{dosovitskiy2017carla,shah2018airsim}.

\begin{figure}[t] \centering
    \includegraphics[width=0.488\textwidth]{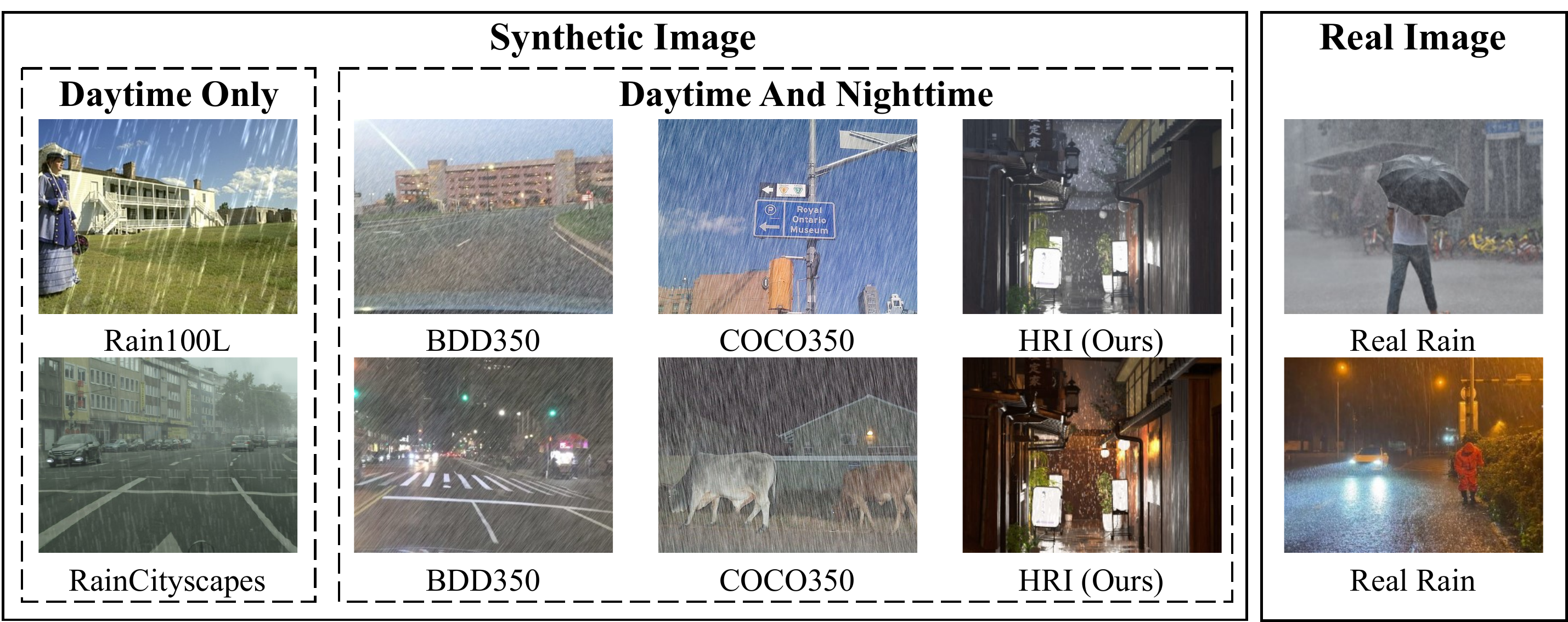}
    \caption{Visual comparisons of real rainy images and synthetic rainy images from four existing datasets and our datasets. Rain100L \cite{yang2017deep} and RainCityscapes \cite{hu2019depth} datasets only contain rainy images in daytime. BDD350 \cite{jiang2020multi} and COCO350 \cite{jiang2020multi} datasets contain rainy images in both daytime and nighttime, but they neglect the influence of scene illumination and depth. Our HRI datasets contain rainy images under complex illumination conditions. Real Rain is captured in real rainy scenes.} 
    \label{fig:figure_rainy_images}
\end{figure}

However, existing synthetic datasets and synthesizers primarily focus on daytime scenes but lack extreme rainy scenes under complex illumination conditions such as nighttime. As presented in Fig. \ref{fig:figure_rainy_images}, Rain100L \cite{yang2017deep} and RainCityscapes \cite{hu2019depth} datasets only contain rainy images in daytime, while BDD350 \cite{jiang2020multi} and COCO350 \cite{jiang2020multi} datasets contain rainy images in both daytime and nighttime but neglect the influence of scene illumination and depth, resulting in significant color appearance differences compared to real rainy images.

Existing rainy image synthesis methods mainly fall into rendering-based methods \cite{garg2006photorealistic,halder2019physics} and learning-based methods \cite{ni2021controlling,ye2021closing,wang2021rain,zhou2024controllable}. Rendering-based methods, which focus on modeling rain streak appearance to render realistic rainy scenes and have advantages in realism, are limited in controllability by complex input data and empirical parameters. Learning-based methods, which train generative models on rainy image datasets to controllably generate rainy images with rain attributes (intensity, direction, etc.) and offer advantages in controllability, typically neglect essential rain attributes like color and optical effects, leading to unrealistic rain color appearances. 
Consequently, the inability of existing rainy image synthesis methods to balance realism and controllability leads to the inadequacy in the data diversity of extreme rainy images, especially in terms of illumination conditions.

To that end, we propose a learning-from-rendering rainy image synthesizer, which combines the benefits of realism and controllability. In the rendering stage, we propose a 3D rainy scene rendering pipeline to render a realistic high-resolution paired rainy-clean images dataset. In the learning stage, the rendered dataset is used to train a \textbf{H}igh-resolution \textbf{R}ainy \textbf{I}mage \textbf{G}eneration Network (HRIGNet), which can capture the illumination information from the clean background images and conveniently generate extreme rainy images under the same illumination conditions. To controllably generate rain streak images for HRIGNet, we introduce the controllable rain image generation network CRIGNet \cite{zhou2024controllable} (which can control rain intensity, direction and potential attributes). 

To validate the effectiveness of our extreme rainy image synthesizer on semantic segmentation task, we require a continuous set of well-labeled extreme rainy images. Given the significant challenges in collecting and labelling real-world extreme rainy images, we turn to CARLA \cite{dosovitskiy2017carla}, an urban driving simulator widely adopted for its open-source code, protocols, digital assets, and flexible configuration of sensor suites and environmental conditions. Specifically, we integrate our proposed synthesizer with the CARLA and develop CARLARain, an extreme rainy street scene simulator which can obtain paired rainy-clean images and labels under complex illumination conditions for extreme rainy scenes.

We perform augmented training with CARLARain on existing semantic segmentation models, and conduct qualitative and quantitative evaluations on synthetic and real rainy street scene datasets. On the synthetic dataset, the accuracy (mIoU) of the models is improved by $5\% - 8\%$. On the real dataset, the segmentation performance of the models is significantly improved under adverse conditions such as insufficient night lighting and obvious interference from rain streaks. Therefore, CARLARain can generate high-quality datasets, which can effectively improve the performance of semantic segmentation models in extreme rainy scenes.

In summary, our work makes the following contributions:
\begin{itemize}
\item  We innovatively propose a learning-from-rendering rainy image synthesizer, which combines the benefits of realism and controllability. In the rendering stage, we propose a 3D rainy scene rendering pipeline to render realistic high-resolution paired rainy-clean images. In the learning stage, we train a \textbf{H}igh-resolution \textbf{R}ainy \textbf{I}mage \textbf{G}eneration Network (HRIGNet) to controllably generate extreme rainy images conditioned on clean images.

\item To construct a continuous set of well-labeled extreme rainy images, we integrated our proposed synthesizer with the CARLA driving simulator and develop CARLARain, an extreme rainy street scene simulator which can obtain paired extreme rainy-clean images and labels under complex illumination conditions. 

\item Qualitative and quantitative experiments validate that CARLARain can effectively improve the accuracy of semantic segmentation models in extreme rainy scenes, with the models’ accuracy (mIoU) improved by $5\% - 8\%$ on the synthetic dataset and significantly enhanced in real extreme rainy scenarios under complex illuminations.

\end{itemize}

\section{Related Work}
\label{section:related}

\begin{table*}[ht]\centering
    \caption{Features of existing rain and street datasets or acquisition methods. (BG: background. SS: semantic segmentation. IS: Instance Segmentation. BBox: bounding box. OF: Optical Fow. Depth: depth map. Pose: camera pose.)}
    \label{tab:related_work}
    
    \begin{tabular}[t]{p{3.4cm}|C{1.9cm}|c|c|C{1.6cm}|C{2.6cm}|C{3.5cm}}
        \hline
        \multirow{2}{*}{name} & \multirow{2}{*}{category} & \multirow{2}{*}{rain type} & \multirow{2}{*}{\shortstack{extreme\\rain}} & \multirow{2}{*}{label} & \multirow{2}{*}{realism} & \multirow{2}{*}{controbility} \\ 
        ~&~&~&~&~&~&~ \\ \hline
        MPID-Real \cite{li2019single} &  \multirow{2}{*}{real rain} & \multirow{2}{*}{real} & $ < 80\%$ & --- & \multirow{2}{*}{---} & \multirow{2}{*}{---} \\ \cline{1-1}\cline{4-4}\cline{5-5}
        RainDS \cite{quan2021removing} & ~ & ~ & $ < 70\%$ & \multirow{11}{*}{BG} & ~ & ~ \\ \cline{1-1}\cline{1-4} \cline{4-4}\cline{6-7}

        \multirow{2}{*}{Photorealistic rendering \cite{garg2006photorealistic}} & \multirow{5}{*}{ \shortstack{rendering-based\\rain}} & \multirow{10}{*}{synthetic} & \multirow{2}{*}{---} & ~ &  realistic optical effects, realistic depth & adjustable illumination and rain intensity \\ \cline{1-1}\cline{4-4}\cline{6-7}
        Physics-based rendering \cite{halder2019physics} & ~ & ~ & $ < 50\%$ & ~ &  realistic optical effects & \multirow{3}{*}{\shortstack{fixed daytime,\\adjustable rain intensity}} \\ \cline{1-1}\cline{4-4}\cline{6-6}
        \multirow{2}{*}{RainCityscapes \cite{hu2019depth}} & ~ & ~ & \multirow{2}{*}{$< 50\%$} & ~ &  realistic optical effects, realistic depth & ~ \\  \cline{1-1}\cline{1-2} \cline{4-4} \cline{6-6} \cline{7-7}
        
        RICNet \cite{ni2021controlling} & \multirow{5}{*}{ \shortstack{learning-based\\rain}}  & ~ &  \multirow{5}{*}{---} & ~ &  \multirow{5}{*}{\shortstack{realistic rain\\streak distribution}} & adjustable rain intensity \\ \cline{1-1}\cline{7-7}
        JRGR \cite{ye2021closing} & ~ & ~ & ~ & ~ & ~ & no  \\ \cline{1-1}\cline{7-7}
        VRGNet \cite{wang2021rain} & ~ & ~ & ~  & ~ & ~ & adjustable latent attributes \\  \cline{1-1}\cline{7-7}
        \multirow{2}{*}{CRIGNet \cite{zhou2024controllable}} & ~ & ~ & ~ & ~ & ~ & adjustable rain intensity, direction and latent attributes \\ \cline{1-1}\cline{1-4} \cline{5-5} \cline{6-6} \cline{7-7}

        Mapillary Vistas \cite{neuhold2017mapillary} & \multirow{3}{*}{real street} & \multirow{3}{*}{real} &  \multirow{3}{*}{$ < 10\%$} & SS, IS &   \multirow{3}{*}{---} & \multirow{3}{*}{---}  \\ \cline{1-1}\cline{5-5}
        nuScenes \cite{caesar2020nuscenes} & ~ & ~ & ~ & 2D/3D BBox & ~ & ~ \\ \cline{1-1}\cline{5-5}
        ACDC \cite{sakaridis2021acdc} & ~ & ~ & ~ & SS & ~ & ~ \\ \cline{1-1}\cline{1-4} \cline{5-5} \cline{6-6} \cline{7-7}
        
        \multirow{2}{*}{AirSim \cite{shah2018airsim}} & \multirow{7}{*}{street simulator} & \multirow{7}{*}{synthetic} &  \multirow{7}{*}{$ 0\%$} & Depth, IS, OF & \multirow{7}{*}{\shortstack{realistic scene,\\unrealistic rain}} & adjustable rain intensity and road wetness intensity \\ \cline{1-1}\cline{5-5} \cline{7-7}
        \multirow{3}{*}{CARLA \cite{dosovitskiy2017carla}} &~&~&~ & \multirow{5}{*}{\shortstack{SS, IS, Depth,\\2D/3D BBox,\\OF, Pose}} & ~ & \multirow{5}{*}{\shortstack{adjustable time, rain intensity,\\ clouds amount, puddles amount,\\ wind intensity and\\camera wetness intensity}} \\
        ~&~&~&~&~&~&~ \\
        ~&~&~&~&~&~&~ \\ \cline{1-1}
        \multirow{2}{*}{SHIFT \cite{sun2022shift}} &~&~&~ & ~ & ~ & ~ \\
        ~&~&~&~&~&~&~ \\ \hline
    \end{tabular}
\end{table*}

\subsection{Rainy Dataset Acquisition}

Real rainy datasets are acquired by capturing rainy images in the real world.  Li \MakeLowercase{\textit{et al.}} \cite{li2019single} proposed a large-scale benchmark consisting of both synthetic and real
world rainy images of various rain types, but the real ones lack paired clean backgrounds. Quan \MakeLowercase{\textit{et al.}} \cite{quan2021removing} mimicked real rainy scenes by spraying water, which can obtain paired rainy-clean images but is time-consuming.

Synthetic rainy datasets are generated with little or no human intervention, which are more efficient. Existing rainy image synthesis methods mainly fall into two categories: rendering-based methods and learning-based methods.

Rendering-based methods focus on modeling the oscillation of raindrops and the appearance of rain streaks. 
Garg and Nayar \cite{garg2006photorealistic} first studied a rain streak appearance model for realistic rain rendering under different illumination. Based on their work, Halder \MakeLowercase{\textit{et al.}} \cite{halder2019physics} proposed a practical, physically-based approach to render realistic rain in images. The RainCityscapes \cite{hu2019depth} dataset was  created by adopting the images in the Cityscapes \cite{cordts2016cityscapes} dataset as clean background images, leveraging the rain streak appearance model. Despite being able to render realistic rain, these methods require users to specify illumination parameters and their physically simulated raindrop distributions fail to match real complexity, leading to poor controllability.

Learning-based methods use generative models to efficiently generate rainy images and exhibit superior controllability. Ni \MakeLowercase{\textit{et al.}} \cite{ni2021controlling} propose a GAN-based \cite{goodfellow2020generative} network for continuous rain intensity control. Ye \MakeLowercase{\textit{et al.}} \cite{ye2021closing} proposed a CycleGAN-based \cite{zhu2017unpaired} framework that jointly learns rain generation and removal. Wang \MakeLowercase{\textit{et al.}} \cite{wang2021rain} proposed a VAE-based \cite{kingma2013auto} model and obtained an interpretable rain generator. Based on their work, Zhou \MakeLowercase{\textit{et al.}} \cite{zhou2024controllable} proposed CRIGNet, a controllable rain image generation network, which can explicitly control rain intensity and direction, and implicitly control other potential rain attributes. However, these methods typically treat the rain layer as a gray-scale layer and disregard other essential rain attributes such as color and optical effects, leading to unrealistic rain color appearances.

To incorporate both controllability and realism into rainy image synthesis, this paper propose a learning-from-rendering rainy image synthesizer.

\subsection{Visual Perception in Autonomous Driving}

In autonomous driving visual perception, some researchers used multiple sensors and manual annotation to construct datasets with rainy images and labels. The Mapillary Vistas Dataset \cite{neuhold2017mapillary} is a large-scale street-level dataset with semantic and instance labels, captured at various weather (sunny, rainy, cloudy, foggy and snowy). The nuScenes dataset \cite{caesar2020nuscenes}, with a full suite of autonomous vehicle sensors, presents diverse scenes in spatial coverage, weather (sunny, rainy, cloudy) and lighting. The ACDC dataset \cite{sakaridis2021acdc}, created for semantic segmentation on adverse visual conditions, includes images of four adverse conditions: fog, nighttime, rain and snow.


Developing deep learning-based visual perception models requires massive annotated data for high performance. However, real data often struggles to generalize to domain shifts and outliers \cite{jaipuria2020deflating}. In contrast, synthetic data can capture the long-tailed distribution of task-specific environmental factors. Many works have demonstrated the effectiveness of synthetic data augmentation in visual perception. Li \MakeLowercase{\textit{et al.}} \cite{li2025improving}, Alhaija \MakeLowercase{\textit{et al.}} \cite{abu2018augmented}, and Zheng \MakeLowercase{\textit{et al.}} \cite{zheng2023robust} have respectively experimented with synthetic data augmentation for autonomous driving tasks such as semantic instance segmentation, object detection, and adverse weather perception.


To efficiently synthesize labeled datasets for multi-task visual perception models, driving simulators based on video games and graphics engines have emerged as powerful tools, leveraging their high-fidelity environment simulation. Shah \MakeLowercase{\textit{et al.}} \cite{shah2018airsim} presented AirSim, an Unreal Engine-based \cite{sanders2016introduction} simulator providing realistic simulations and a high-frequency physics engine for real-time HITL simulations. Similarly, Dosovitskiy \MakeLowercase{\textit{et al.}} \cite{dosovitskiy2017carla} proposed CARLA, an open-source simulator for autonomous driving research that supports flexible specification of sensor suites and environmental conditions. Based on CARLA, Sun \MakeLowercase{\textit{et al.}} \cite{sun2022shift} created SHIFT, a multi-task synthetic dataset with discrete and continuous shifts in cloudiness, rain/fog intensity, time of day, and vehicle/pedestrian density. However, these existing simulators mainly focus on daytime scenes and lack realistic simulations of extreme rainy scenes under complex illumination conditions.

By contrast, CARLA stands out for its open-source code, protocols, and digital assets, coupled with flexible configuration of sensor suites and environmental conditions, which empower it to synthesize comprehensive label data and simulate diverse scenarios. Thus, to construct a continuous set of well-labeled extreme rainy images, we integrate the proposed synthesizer with CARLA and develop CARLARain, an extreme rainy street scene simulator.

Features of existing rain and street datasets or acquisition methods are summarized in Table \ref{tab:related_work}, including rain types (real or synthetic), the proportion of extreme rainy images, label types, realism and controllability of synthetic rainy images.

\section{  Learning-from-rendering Rainy Image Synthesizer}
\label{section:hrig}

\begin{figure*}[ht] \centering
    \includegraphics[width=0.9\textwidth]{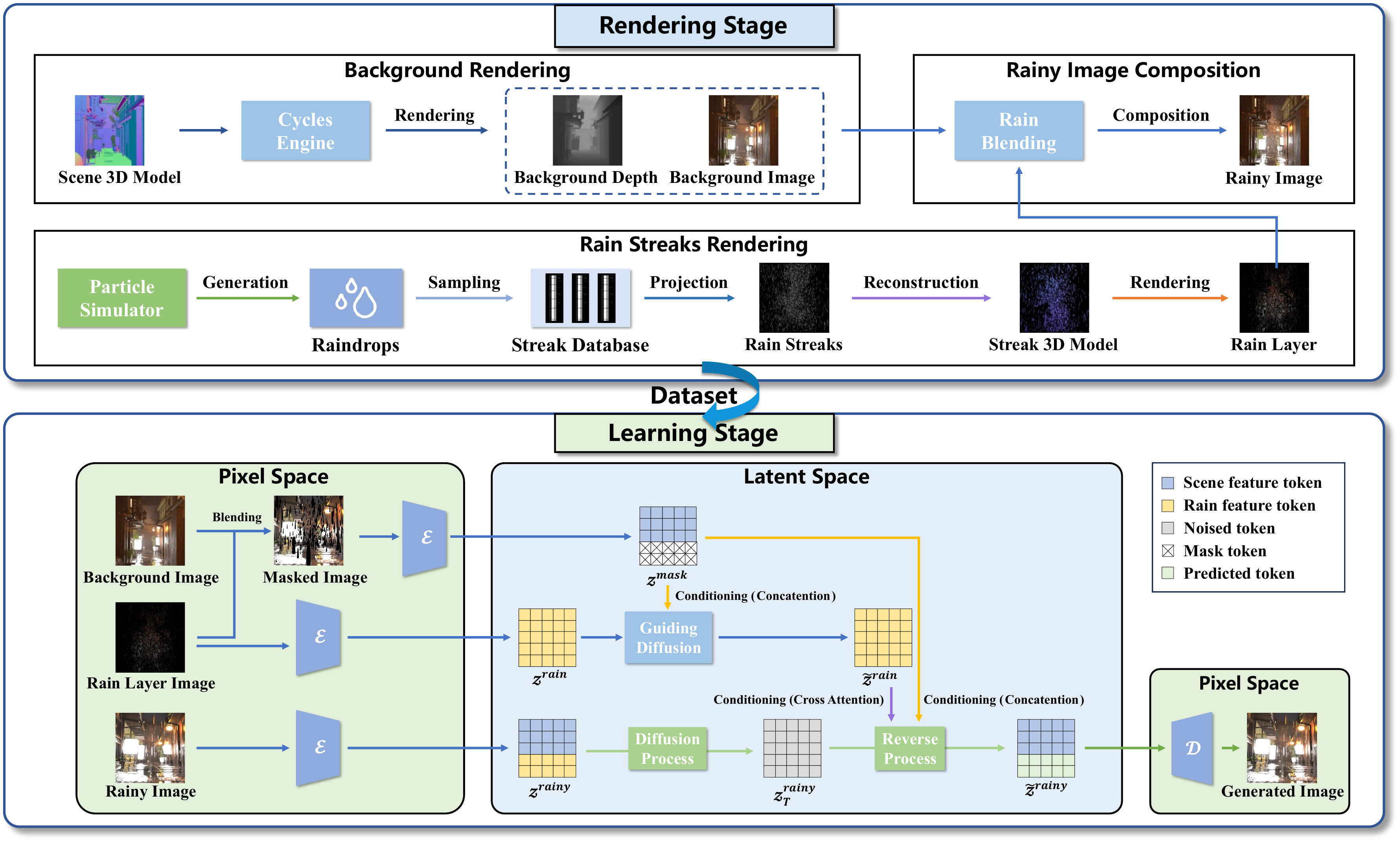}
    \caption{Overview of our learning-from-rendering rainy image synthesizer. The top is the rendering stage, the bottom is the learning stage, and the two stages are interlinked via rainy image datasets.} 
    \label{fig:figure_hrig_pipeline}
\end{figure*}

In extreme rainfalls, images captured by cameras are disturbed by rain in the scene and suffer from degradation such as rain streaks, raindrops, fog-like rain, etc. In this section, we mainly focus on the phenomenon of rain streaks, which occurs when falling raindrops produce motion-blurred streaks during the exposure time of the camera. Rainy images in this paper refer to images that have been degraded by rain streaks. 

As mentioned above, we propose a learning-from-rendering rainy image synthesizer, as shown in Fig. \ref{fig:figure_hrig_pipeline}. Specifically, the synthesizer combines rendering-based and learning-based methods in two stages: the rendering stage and the learning stage. In the rendering stage, we propose a 3D rainy scene rendering pipeline to render realistic high-resolution paired rainy-clean images and create a high-resolution paired rainy-clean image dataset. In the learning stage, we train a rainy image generation network with the rendered datasets to conveniently generate rainy images under the same illumination conditions, conditioned on clean background images and rain streak images.

To further validate the effectiveness of our extreme rainy image synthesizer on semantic segmentation task, we integrated the synthesizer with the CARLA \cite{dosovitskiy2017carla} driving simulator and develop CARLARain, an extreme rainy street scene simulator which can obtain paired extreme rainy-clean images and labels under complex illumination conditions.

\subsection{Rendering Stage}

To train a high-quality rainy image generation network, realistic paired rainy-clean images under diverse illumination are essential. Given that offline rendering techniques based on ray tracing algorithm \cite{pharr2016physically} can simulate most natural phenomena of object surface interactions and produce realistic images, they are ideal for this task. Currently, these techniques have found extensive use in the fields of movies, animation, and design. Blender \cite{blender2018}, a popular open-source 3D creation suite, has a powerful built-in unbiased path-tracer engine named Cycles, which can render realistic scene images.

In the rendering stage, we propose a 3D rainy scene rendering pipeline integrating Blender and the image-based rain rendering algorithm \cite{garg2006photorealistic,halder2019physics}. As shown in Fig. \ref{fig:figure_hrig_pipeline}, it can render realistic paired rainy-background images under any illumination conditions based on the scene 3D model. 
Specifically, we create 3D scene models and render them with Cycles engine to obtain background RGB images and depth images. 
Meanwhile, we implement a raindrop physical particle simulator to generate raindrop particles, with invisible ones excluded to reduce the amount of rendering data.
Furthermore, using the rain streak rendering algorithm proposed by Garg and Nayar \cite{garg2006photorealistic}, we sample and project rain streaks from the rain streak database into the image.
Then, we reconstruct each rain streak pixel into a quad to get rain streak 3D models. After merging them with the background scene 3D model, we render the resultant rainy scene 3D model for illuminated rain layers. 
Finally, we blend rain layers with background images to get rainy images. Each image pair thus comprises a background, a depth, a rain layer, and a rainy image.

\textbf{Background rendering.} Using the Cycles engine, we render the pre-created scene 3D model for realistic background RGB and depth images. To capture full-day scenes under various illumination, we employ a procedural sky and set equal time intervals from 0 to 24 hours. For camera settings, we used focal lengths of 30-50mm, and a 1/60s shutter speed 
 \cite{garg2006photorealistic}.

\textbf{Rain streaks rendering.} To match real-world raindrop size and distribution, we implement a raindrop particle physics simulator, referring to existing raindrop dynamics studies \cite{duhanyan2011below,best1950size,kessler1969distribution}. The simulation involves raindrop size, terminal velocity, and spatial distribution.\footnote{More derivations are included in supplementary material.}


To reduce rendering data, we cull invisible raindrops generated by our simulator. First, we remove raindrops outside the camera frustum. Then, using depth map and raindrop positions, we further cull occluded raindrops. Finally, we sample rain streaks from the database \cite{garg2006photorealistic} and project onto the image following the image-based rain rendering algorithm \cite{halder2019physics}.

In the image-based rain rendering algorithm \cite{garg2006photorealistic}, raindrops require scene light source illumination. To match the background scene's illumination, we generate rain streak 3D models by creating a quad at each raindrop pixel. After merging these models with the scene 3D model, we render them under the same illumination conditions as the background to obtain an illuminated rain layer.

\textbf{Rainy image composition.} We composite the background and illuminated rain layer images to generate rainy images. According to the image-based rain rendering algorithm  \cite{halder2019physics}, assuming that $\mathbf{x}$ is a pixel in the background image $I$ and $\mathbf{x}'$ is the overlapping coordinate in rain layer $S'$, the blending formula is:
\begin{equation}
    I_{rainy}(\mathbf{x})=\frac{T-S'_{\alpha}(\mathbf{x'})\tau_1}{T}I(\mathbf{x})+S'(\mathbf{x'})\frac{\tau_1}{\tau_0},
\end{equation}
where $S'_{\alpha}(\mathbf{x'})$ is the alpha channel of the rain layer, $T$ is the targeted exposure time, $\tau_0=\sqrt{10^{-3}}/50$ is the time for which the raindrop remained on one pixel in the streak database, and $\tau_1$ is the same measure according to our physical simulator.

\subsection{Learning Stage}


As shown in Table \ref{tab:avg_resolution}, we statistic the average image resolution of five synthetic rainy image datasets. In recent years, visual perception models have demanded increasingly higher image resolutions. For instance, Cityscapes \cite{cordts2016cityscapes}—one of the most commonly used street scene datasets—has a resolution of $2048 \times 1024$, yet existing synthetic rainy image datasets often fail to meet this requirement. Thus, in the learning stage, designing a network for high-resolution rainy image generation is crucial, as it enables the generated rainy images to match the input resolution of existing visual perception models.

\begin{table}\centering
    \caption{Average resolution of five synthetic rainy image datasets.}
    \begin{tabular}{cc}
        \hline
        dataset & image resolution \\
        \hline
        Rain100L \cite{yang2017deep} & $400\times400$ \\
        COCO350 \cite{jiang2020multi} & $640\times480$ \\
        BDD350 \cite{jiang2020multi} & $1280\times720$ \\
        RainCityscapes \cite{hu2019depth} & $2048\times1024$ \\
        HRI (Ours) & $2048\times1024$ \\
        \hline
    \end{tabular}
    \label{tab:avg_resolution}
\end{table}

To this end, in the learning stage, we propose a High-resolution Rainy Image Generation Network (HRIGNet), which generates high-resolution rainy images from clean background images and corresponding rain layer masks. Specifically, given an RGB background image and a mask indicating rain streak positions, HRIGNet can generate rain streaks with illumination and color consistent with the background at corresponding positions.

As shown in Fig. \ref{fig:figure_hrig_pipeline}, the architecture of HRIGNet is based on the LDM \cite{rombach2022high}, where an autoencoding model \cite{esser2021taming} is employed to perceptually compress images into latent code and the forward and reverse processes of diffusion models are conducted in the latent space.

To control the generation process of diffusion models, we adopt concatenation and cross-attention as conditional mechanisms for the reverse process. Specifically, we use a guiding diffusion model to predict the latent code of rain layer images, which enhances the underlying UNet backbone \cite{ronneberger2015u} of diffusion models via cross-attention. This provides more guidance information to improve generated rainy images' quality. To impose stronger generation constraints, we merge the background and rain layer mask into a masked image, using it as a condition for the reverse process via concatenation.

\textbf{Latent Diffusion model.} Diffusion models\cite{rombach2022high,peebles2023scalable} are probabilistic models that gradually add noise to the data by traversing a Markov chain of $T$ time steps, transforming the distribution of real data to a Gaussian distribution. To reduce the training cost of high-resolution diffusion models, LDM \cite{rombach2022high} (Latent Diffusion Model) use a pretrained VQGAN \cite{esser2021taming} autoencoding model to compress images from the high-dimensional image space to a low-dimensional latent space. This enables more efficient diffusion model training in the latent space.
Given a high-resolution RGB rainy image $x^{rainy}$, the corresponding latent code encoded by the encoder $\mathcal{E}$ is $z^{rainy}$. So the objective of LDM can be expressed as: 
\begin{equation}
L_{LDM}=\mathbb{E}_{\mathcal{E}(x^{rainy}),\epsilon\sim \mathcal{N}(0,1),t} \Big[\left \|  \epsilon - \epsilon_{\theta}(z_t^{rainy},t) \right \|_2^2 \Big],
\end{equation}
where timestep $t$ is uniformly sampled from $\{ 1,\dots,T \}$, $z_t^{rainy}$ is a noisy version of the input $z^{rainy}$ at timestep $t$, $\epsilon$ represents the noise sampled from a standard normal distribution $\mathcal{N}(0,1)$, and $\epsilon_{\theta}(z^{rainy}, t)$ is the denoising network's prediction of the noise component in the input data, with $\theta$ being the parameters of the model. Here, the neural backbone $\epsilon_{\theta}(\circ,t)$ of LDM is realized as a UNet \cite{ronneberger2015u}.

\textbf{Guiding Diffusion model.} Guiding high-resolution image generation with intermediate results provides additional information, which is expected to improve the quality of generated images. Therefore, we use a guiding diffusion model (GDM) trained on the latent codes of rain layer images. The GDM coarsely predicts the latent codes of the rain layer images, which are used as a condition in the reverse process of the diffusion model. 

Specifically, we first merge the input RGB background and the rain layer mask into a masked image $x^{mask}$. Then the masked image is encoded into the latent space to obtain the latent codes $z^{mask}$, which are input to the GDM to predict the latent code of rain layer image $\tilde{z}^{rain}$. The objective of GDM can be expressed as:\par

\begin{small}
\begin{equation}
L_{GDM}= \mathbb{E}_{\mathcal{E}(x^{rain}),\epsilon\sim \mathcal{N}(0,1),t} \Big [\left \|  \epsilon - \epsilon_{\theta}(z_t^{rain},t,z^{mask}) \right \|_2^2 \Big ],
\label{equ:loss_gdm}
\end{equation}
\end{small}%
where $x^{rain}$ is the high-resolution RGB rain layer image, and the corresponding latent code encoded by the encoder is $z^{rain}$. Here, the neural backbone of GDM is realized as a UNet \cite{ronneberger2015u}.

\textbf{Conditioning mechanisms.} In the context of image generation, LDMs use the cross-attention mechanism \cite{vaswani2017attention} to enable inputs from different modalities to serve as conditionings. Our method leverages the predicted latent code of the rain layer image by GDM as a condition via cross-attention. Specifically, $\tilde{z}^{rain}$ is mapped to the intermediate layers of the UNet via a cross-attention layer represented as Attention$(Q,K,V)=$softmax$(\frac{QK^T}{\sqrt{d}})\cdot V$, where 
$$Q=W^{(i)}_Q\cdot \varphi_i(z_t^{rainy}) ,K=W^{(i)}_K\cdot \tilde{z}^{rain},V=W^{(i)}_V\cdot \tilde{z}^{rain}.$$ 
Here, $\varphi_i(z_t^{rainy})$ represents the intermediate representation of the denoising network $\epsilon_{\theta}$ implemented by UNet, and $W^{(i)}_Q, W^{(i)}_K, W^{(i)}_V$ are learnable projection matrices.

Moreover, to enhance the constraints on the image generation process, we combine the concatenation and cross-attention conditioning mechanisms. We use the latent code of masked image $z^{mask}$ as a condition of the reverse process via concatenation. Specifically, the input of the reverse process are $z_t^{concat} = [z^{rainy}_t, z^{mask}]$. 

Via the concatenation and cross-attention conditioning mechanisms, we then learn the conditional LDM via:\par
\begin{small}
\begin{equation}
L_{LDM}=\mathbb{E}_{\mathcal{E}(x^{rainy}),\epsilon\sim \mathcal{N}(0,1),t} [\left \|  \epsilon - \epsilon_{\theta}(z_t^{concat},t,\tilde{z}^{rain}) \right \|_2^2].
\label{equ:loss_ldm}
\end{equation}
\end{small}%

Combining the two objective functions in Eq. \ref{equ:loss_gdm} and Eq. \ref{equ:loss_ldm}, the total objective of HRIGNet is:
\begin{equation}
L_{HRIG} = L_{GDM} + L_{LDM}
\end{equation}

\textbf{Image generation.} In LDM, for images of size $H\times W$ with $m$ downsampling blocks, the input latent code to the diffusion model is of size $H/2^m \times W/2^m$, which reduces the spatial cost of training and speeds up training and inference processes. However, when $m$ exceeds a critical value, reconstruction quality degrades, as demonstrated in \cite{esser2021taming}. Therefore, during training, we have to work patch-wise and crop images. Specifically, we train our model using images of up to $512\times512$ resolution.
Considering that spatial conditioning information \cite{esser2021taming} is available in our model, during inference, we can simply segment background images into patches of size $512\times512$ and use them as conditioning. By merging the results, we can generate images up to $2048\times1024$ resolution.

In our implementation, simply segmenting background images into blocks and feeding them to the model leads to hue variations in output results, causing a blocky effect in the merged output. We address this by using rain layer masks to blend generated images and background images.

\subsection{Extreme Rainy Street Scene Simulator}

\begin{figure}[ht] \centering
    \includegraphics[width=0.488\textwidth]{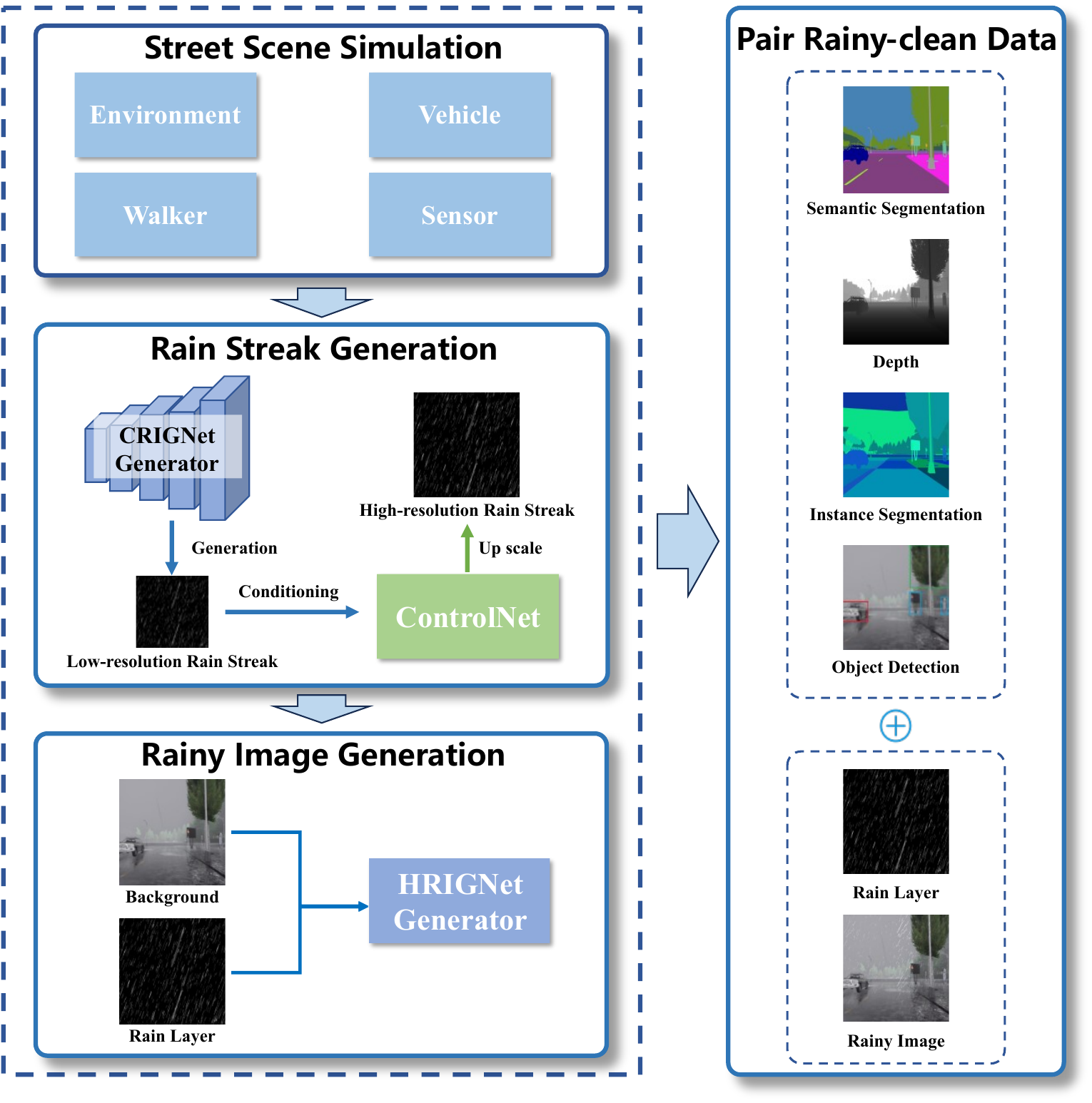}
    \caption{Overview of our extreme rainy street scene simulator, CARLARain. By integrating the proposed HRIGNet with the CRIGNet and the CARLA, CARLARain can obtain paired extreme rainy-clean images and label information data under complex illumination conditions.} 
    \label{fig:figure_carlarain}
\end{figure}

This paper integrates our proposed HRIGNet with the CARLA \cite{dosovitskiy2017carla} driving simulator and the controllable rain image generation network CRIGNet \cite{zhou2024controllable}, constructing CARLARain, an extreme rainy street scene simulator, as shown in Fig. \ref{fig:figure_carlarain}. Next, we will introduce the implementation of CARLARain in detail.




\textbf{Street scene simulation.}
To enable realistic physical and visual street scene simulations and obtain labeled data for visual perception models, we introduce the CARLA driving simulator. CARLA \cite{dosovitskiy2017carla} is an open-source urban driving simulator implemented as an open-source layer over Unreal Engine (UE) \cite{lee2024unreal}, which provides state-of-the-art rendering quality \cite{lee2024unreal} and realistic physics. CARLA allows for flexible configuration of commonly used visual sensors, enabling the collection of scene images and label data during simulation. In this paper, our CARLARain configures RGB camera, depth camera, semantic segmentation camera, instance segmentation camera, and collision detection sensor based on CARLA. During simulation, CARLARain can capture frame-by-frame street scene data, RGB, semantic/instance segmentation and depth images, as well as object bounding boxes. 




\textbf{Rain streak generation.} 
Due to the limitations of the Variational Autoencoder (VAE) on which CRIGNet is built, it can only generate rain streak images with a maximum resolution of $256 \times 256$.  
To generate high-resolution rain streak images, we pretrain a ControlNet-based \cite{zhang2023adding} up-scale model.\footnote{Details are included in supplementary material.}
Using low-resolution rain streak images from CRIGNet as control conditions, it generates corresponding high-resolution ones. By inputting rain intensity and direction variables into CRIGNet and ControlNet, CARLARain can controllably generate rain streak images for rainy image generation.

\textbf{Rainy image generation.} After acquiring clean scene RGB images and rain streak images, CARLARain can generate realistic rainy images under the same illumination conditions as the background environment by inputting these images as conditional information into the HRIGNet. This allows CARLARain to produce paired extreme rainy-clean images and label data under complex illumination conditions.

\section{Experimental Results}
\label{section:experiment}

In this section, we conduct qualitative and quantitative experiments to validate the realism and controllability of our proposed rainy image synthesizer, as well as its effectiveness in improving the performance of semantic segmentation models in extreme rainy scenes. \footnote{Implementation details are included in supplementary material.}

For rainy image generation, we compare HRIGNet with some baseline image generative models to validate its performance in high-resolution rainy image generation. Moreover, we conduct image deraining experiments to validate that it can improve the accuracy of image deraining models, thus indirectly validating the realism of the generated rainy images. Additionally, we demonstrate some rainy images, illustrating their controllability in attributes such as illumination, rain intensity, and direction.

Based on CARLARain, we construct an extreme rainy street scene image dataset, ExtremeRain. We test existing semantic segmentation models on it to evaluate the impact of rainy scenes under different illumination conditions. Furthermore, to improve model accuracy in extreme rainy scenes, we conduct augmented training with ExtremeRain and perform qualitative and quantitative evaluations on synthetic and real datasets.

\subsection{Dataset}

\textbf{High-resolution Rainy Image dataset.} To train the HRIGNet, we create a High-resolution Rainy Image (HRI) dataset in the rendering stage of the proposed rainy image synthesizer. The HRI dataset comprises a total of 3,200 image pairs. Each image pair comprises a clean background image, a depth image, a rain layer mask image, and a rainy image. As shown in Table \ref{tab:hri_dataset}, it contains three scenes: lane, citystreet and japanesestreet, with image resolutions of $2048\times1024$.\footnote{Details about HRI dataset are included in  supplementary material.}

\begin{table}\centering
    \caption{Overview of the HRI dataset.}
    \begin{tabular}{cccccc}
        \hline
        scene & 
        \makecell[c]{ dataset \\type}& 
        camera &
        frames & 
        intensities& 
        \makecell[c]{image \\ pairs}\\ 
        \hline
        \multirow{2}{*}{lane} & train set & 
        3 & 
        \multirow{2}{*}{100} & 
        \multirow{2}{*}{4} & 1,200 \\ 
        ~ & test set & 1 & ~ & ~ & 400 \\ 
       \multirow{2}{*}{citystreet} & train set &
       5 & 
       \multirow{2}{*}{25} & 
       \multirow{2}{*}{4} & 500 \\ 
        ~ & test set& 1 & ~ & ~ & 100 \\ 
       \multirow{2}{*}{japanesestreet} & train set &
       8 & 
       \multirow{2}{*}{25} & 
       \multirow{2}{*}{4} & 800 \\ 
        ~ & test set & 2 & ~ & ~ & 200 \\ 
        \hline
    \end{tabular}
    \label{tab:hri_dataset}

\end{table}

\textbf{ExtremeRain dataset.} As shown in Table \ref{tab:extreme_rain}, the ExtremeRain dataset contains 8 different scenes and 3 illumination conditions: daytime, sunset, night. The rainy scenes feature a rain intensity ranging from $5 mm/h - 100 mm/h$, covering extreme rainfalls under complex illumination conditions. It contains labels for semantic segmentation, instance segmentation, depth estimation, and object detection.\footnote{Details about ExtremeRain dataset are included in supplementary material.}

\begin{table}\centering
    \caption{Overview of the ExtremeRain dataset.}
    \begin{tabular}{ccccccc}
        \hline
        \makecell[c]{ dataset type}& 
        scene &
        time &
        frames & 
        samples \\
        \hline
        train set & 7 & 3 & 1000 & 21000 \\ 
        test set & 1 & 3 & 1000 & 3000  \\ 
        \hline
    \end{tabular}
    \label{tab:extreme_rain}
\end{table}

\subsection{Realism And Controllability Of Rainy Image Generation}

\begin{table}[t]\centering
    \caption{Quantitative evaluation of baseline models and HRIGNet.}
    \resizebox*{0.48\textwidth}{!}{
        \begin{tabular}{cccccc}
            \hline
            method & patch size & FID$\downarrow$ & LPIPS$\downarrow$ & SSIM$\uparrow$ & PSNR$\uparrow$ \\ 
            \hline
            LDM \cite{rombach2022high} & 512$\times$512 & 46.583 & 0.241 & 0.703 & 16.652  \\ 
            DiT \cite{peebles2023scalable} & 512$\times$512 & 164.977 & 0.490 & 0.548 & 12.097\\ 
            CycleGAN \cite{zhu2017unpaired} & 512$\times$512 &  47.073 & 0.271  & 0.639  & \textbf{21.604}
            \\
            HRIGNet & 512$\times$512 & \textbf{32.111} & \textbf{0.205} & \textbf{0.747} & 18.595  \\ 
            \hline
        \end{tabular}
    }
    \label{tab:table_baseline}
\end{table}

\begin{figure}[t] \centering
    \captionsetup[subfloat]{labelfont=scriptsize,textfont=scriptsize}
    \subfloat[Background]{
        \includegraphics[width=0.15\textwidth]{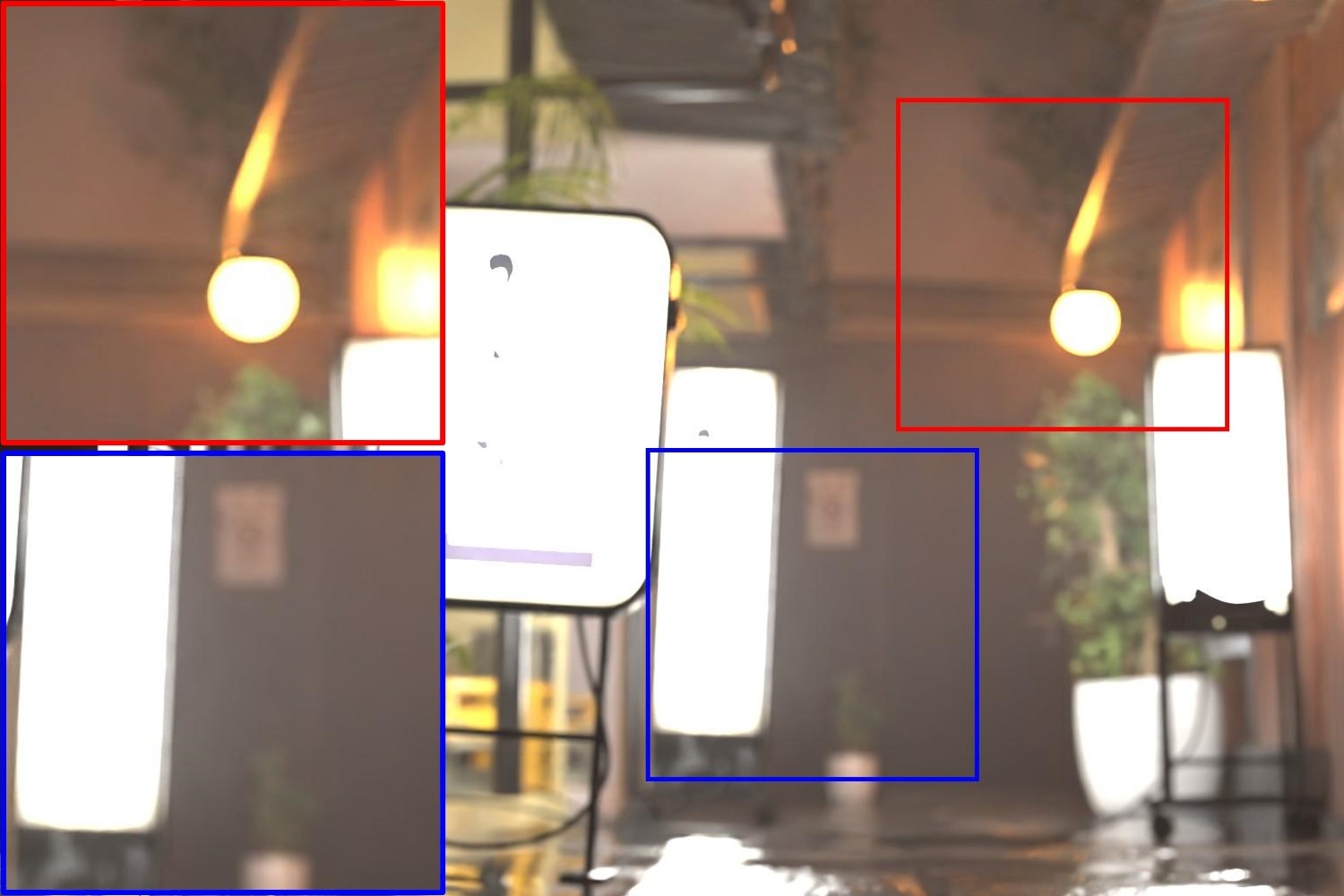}
    }
    \subfloat[LDM \cite{rombach2022high}]{
        \includegraphics[width=0.15\textwidth]{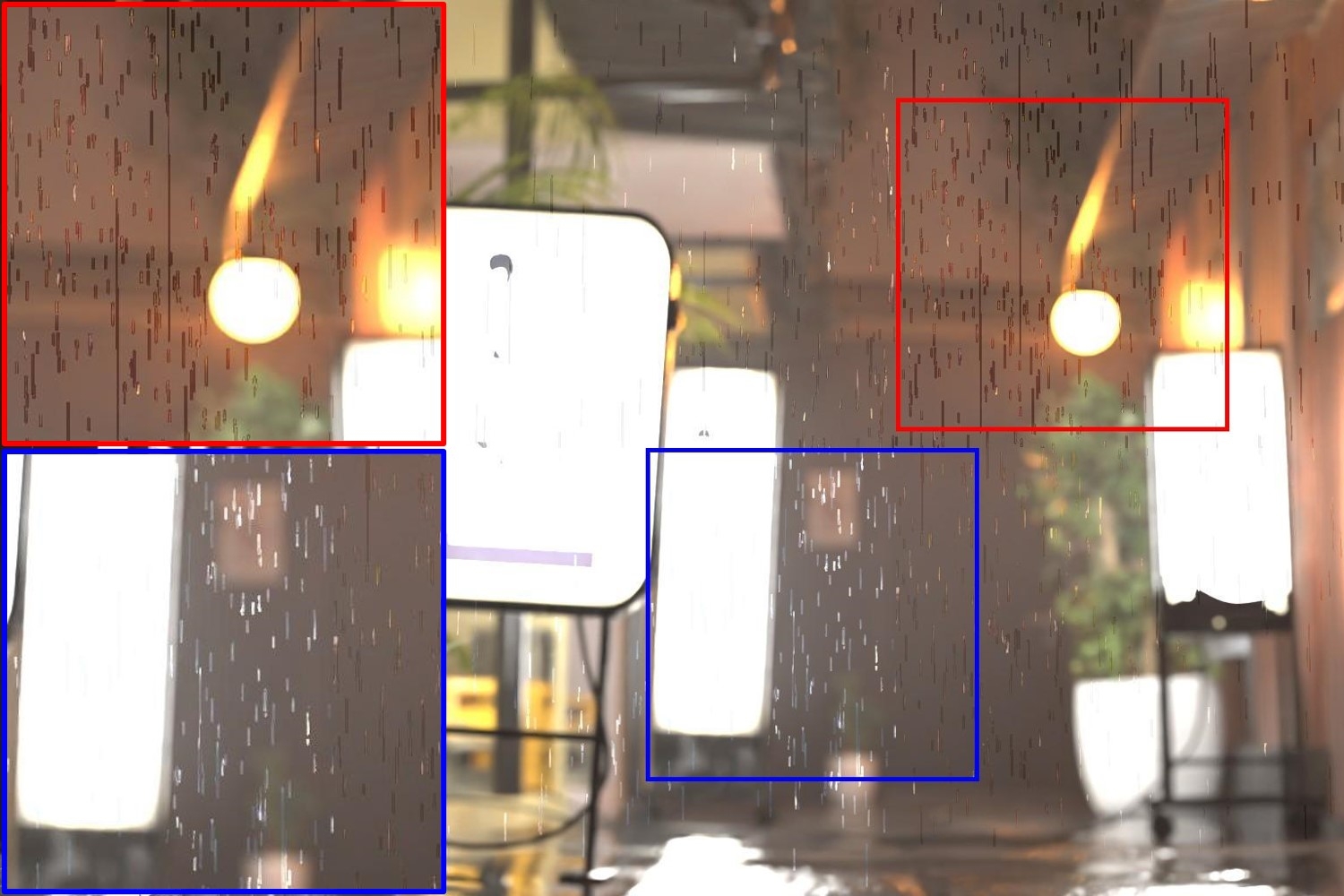}
    }
    \subfloat[DiT \cite{peebles2023scalable}]{
        \includegraphics[width=0.15\textwidth]{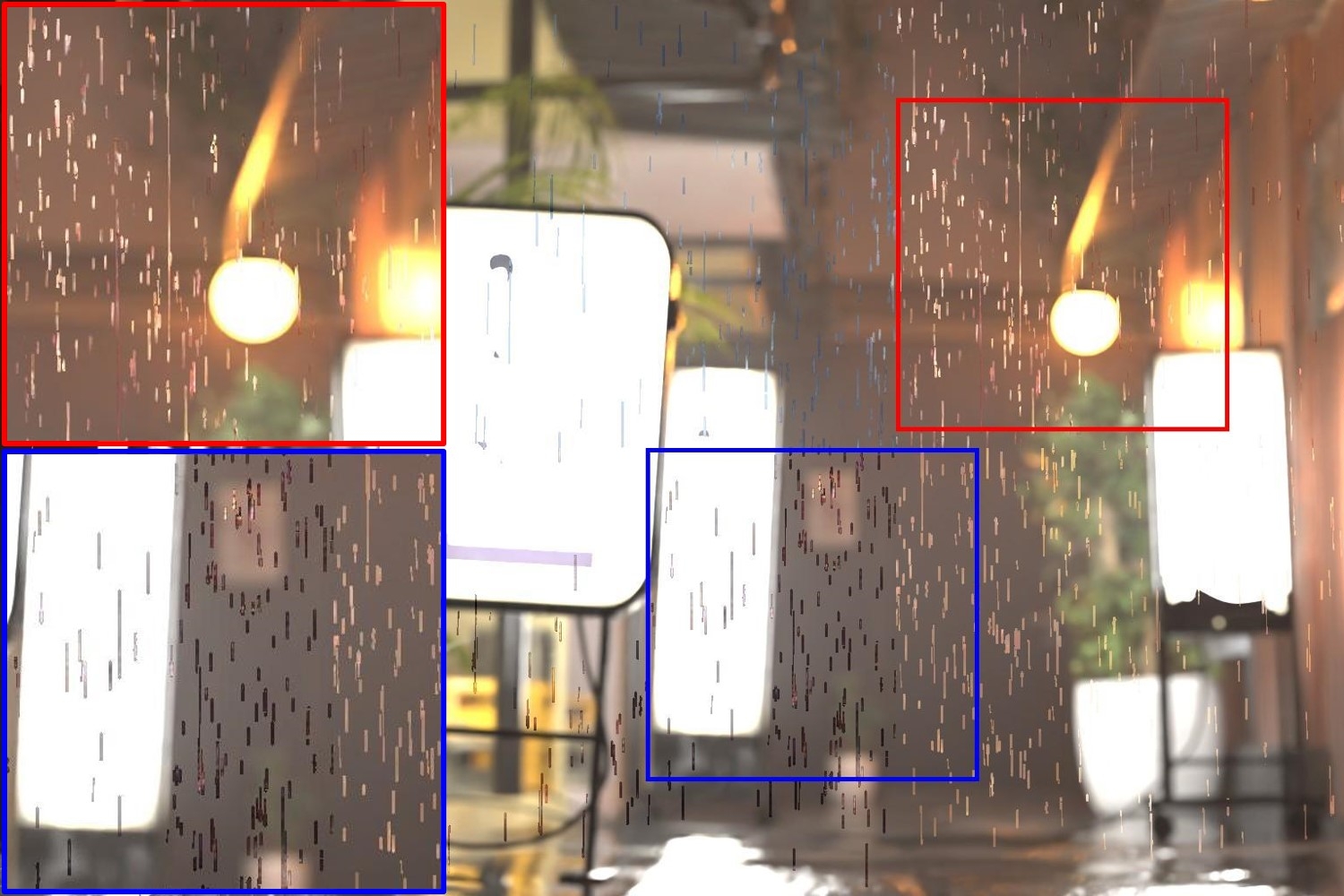}
    }
    \\
    \subfloat[Ground truth]{
        \includegraphics[width=0.15\textwidth]{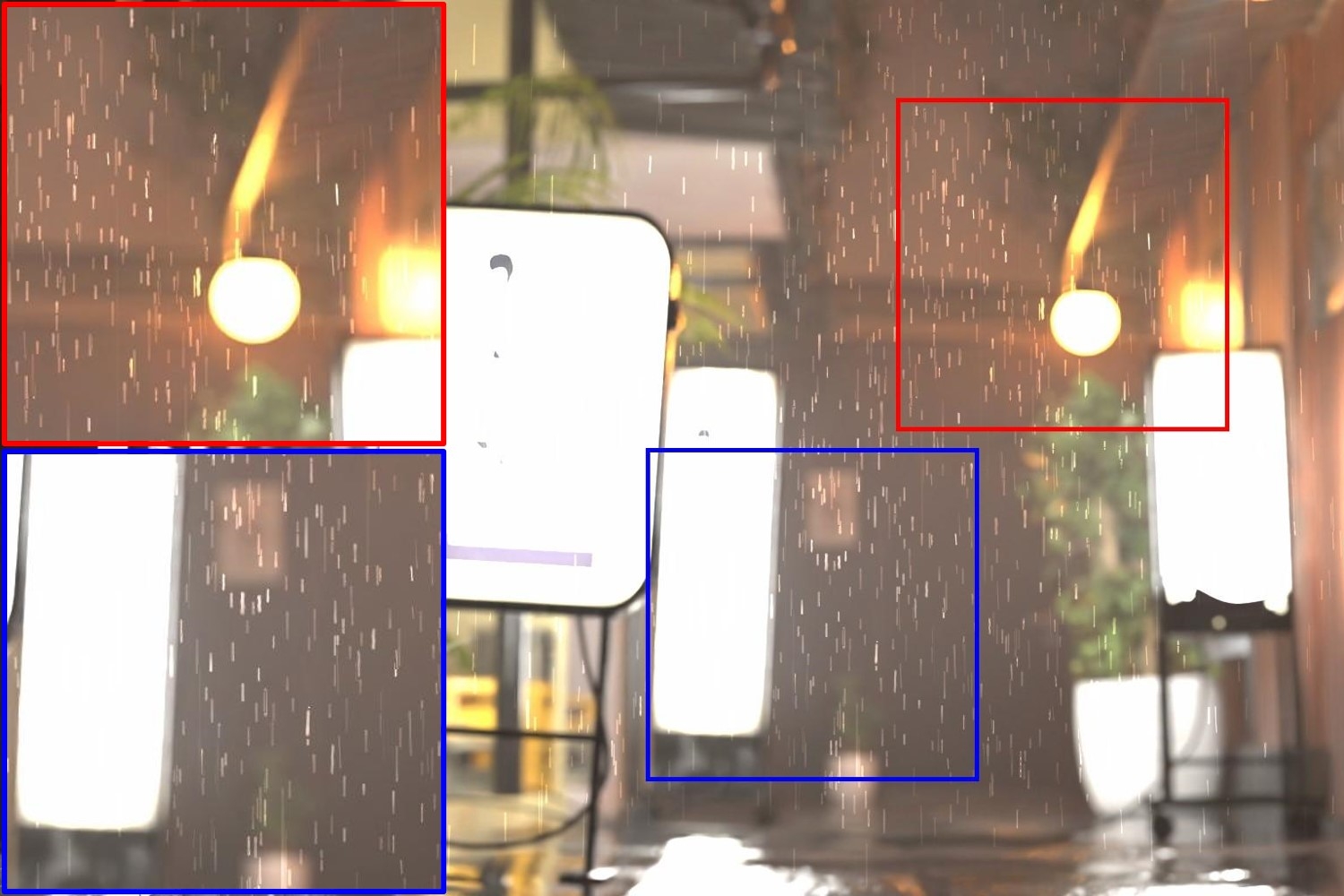}
    }
    \subfloat[CycleGAN \cite{zhu2017unpaired}]{
        \includegraphics[width=0.15\textwidth]{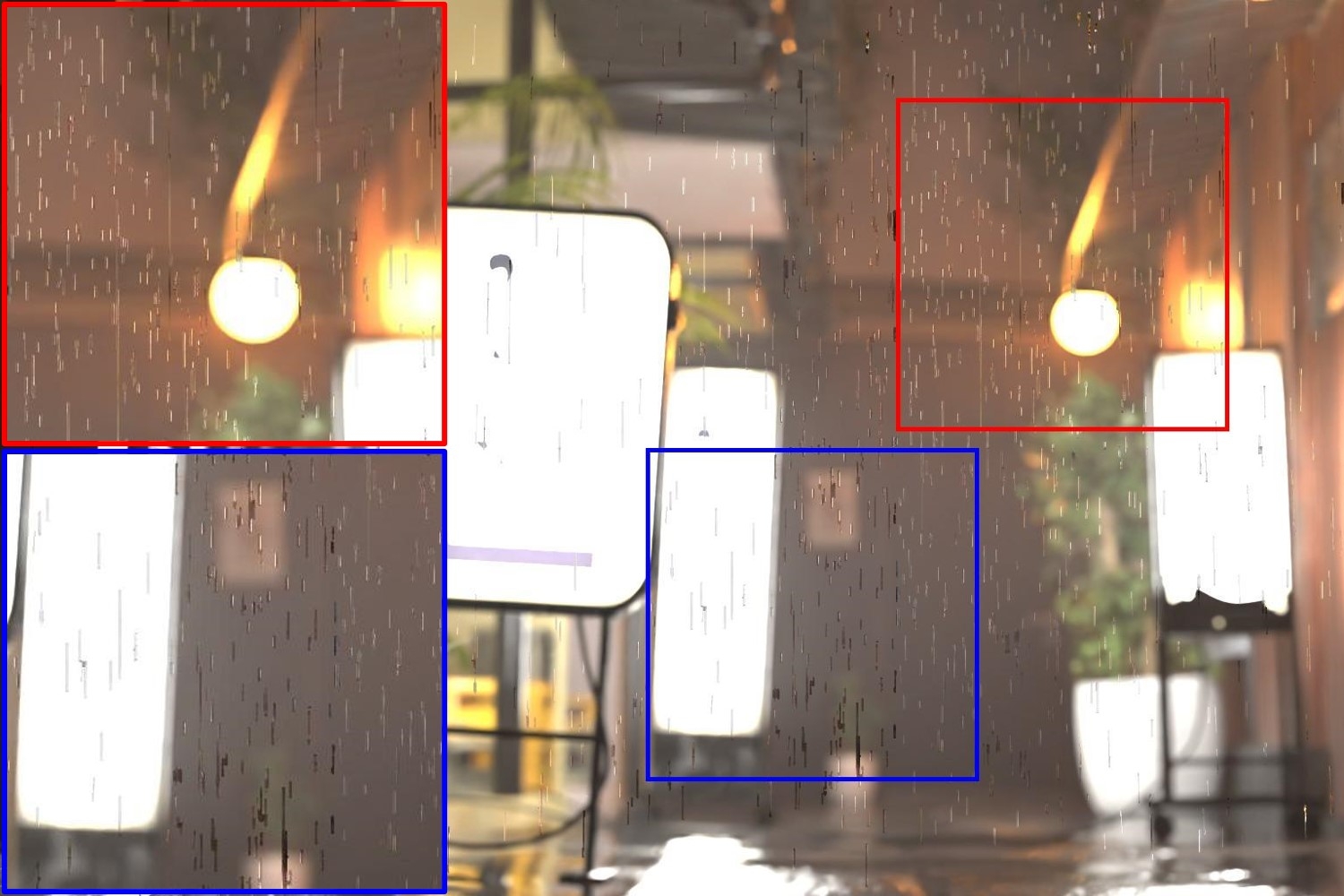}
    }
    \subfloat[HRIGNet (ours)]{
        \includegraphics[width=0.15\textwidth]{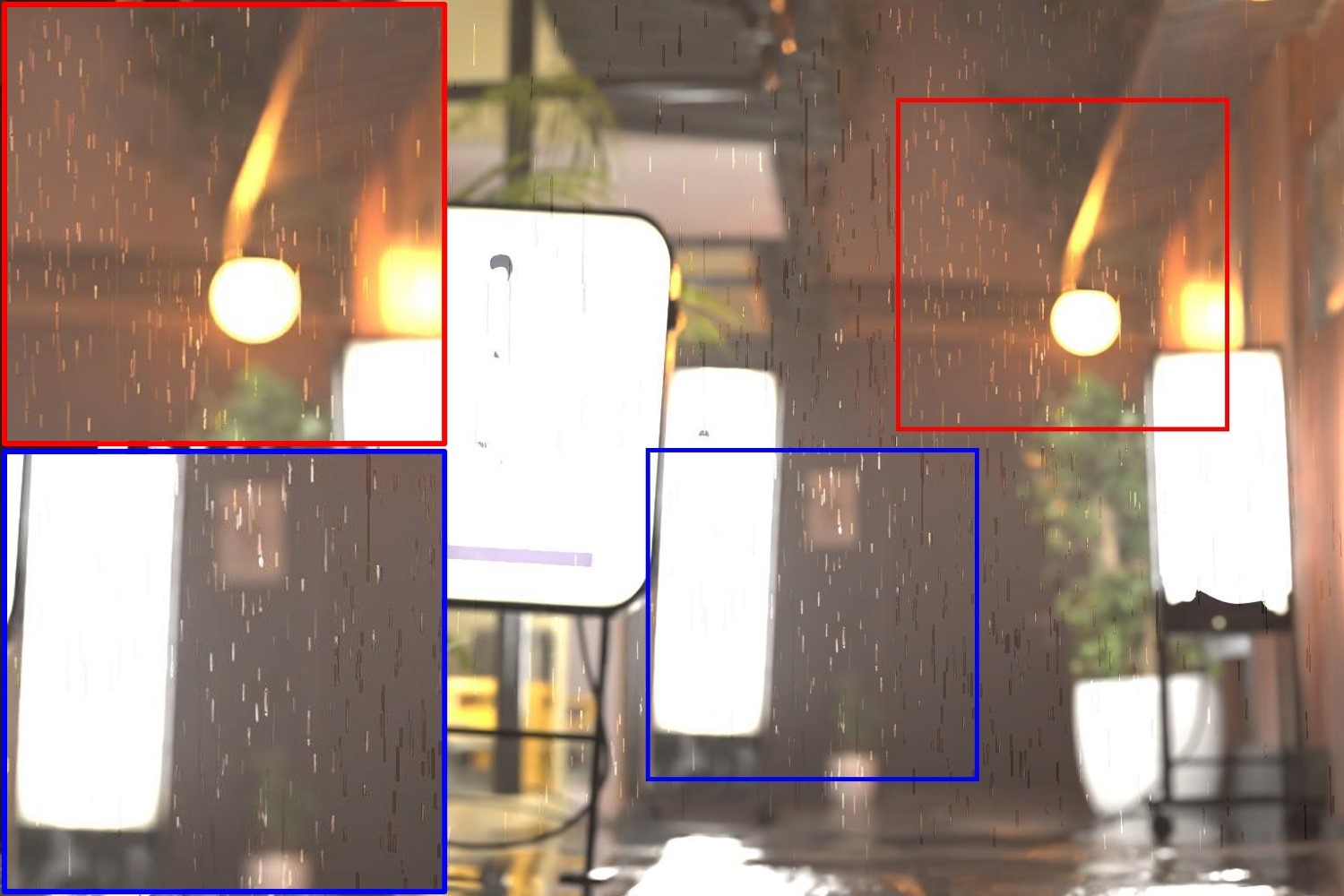}
    }

    \caption{Visual comparison of generated rainy images from different image generative models. Our HRIGNet can better capture the illumination and color in the background image, and map them to the generated rain layer.} 
    \label{fig:figure_baseline}
\end{figure}



\textbf{Compare with baseline.} To evaluate the performance of HRIGNet in high-resolution rainy image generation, we compare it with several baseline image generative models: LDM \cite{rombach2022high}, DiT \cite{peebles2023scalable} and CycleGAN \cite{zhu2017unpaired}. The evaluation metrics used are FID \cite{heusel2017gans}, LPIPS \cite{zhang2018unreasonable}, SSIM, and PSNR. As the results shown in Table \ref{tab:table_baseline}, our model achieves state-of-the-art results in FID, LPIPS and SSIM. Fig. \ref{fig:figure_baseline} illustrates a comparison of rainy image generation results of these methods. As seen, our method can well capture the illumination and color in the background image, and map them to the generated rain layer. As a result, the rain layer is endowed with a visually plausible color appearance that matches the background image.

\begin{table}[t]\centering
    \centering
    \caption{Quantitative comparisons on test data of SPA-Data.}
    \resizebox*{0.49\textwidth}{!}{
        \begin{tabular}{c|ccc|ccc}
        \hline
           dataset & \multicolumn{3}{c|}{RainTrainL} & \multicolumn{3}{c}{Rain1400} \\ \hline
           metric & FID$\downarrow$ & SSIM$\uparrow$ & PSNR$\uparrow$ & FID$\downarrow$ & SSIM$\uparrow$ & PSNR$\uparrow$  \\ \hline
            PReNet & 45.981 & 0.941 & 33.493 & 48.834 & 0.937 & 31.869   \\ 
            PReNet+ & 45.586 & 0.942 & 33.168 & 49.001 & 0.934 & 32.157    \\ 
            PReNet++ & \textbf{43.706} & \textbf{0.946} & \textbf{33.614} & \textbf{46.073} & \textbf{0.944} & \textbf{33.064}   \\
            $\Delta\uparrow$ & 2.275 & 0.005 & 0.121 & 2.761 & 0.007 & 1.195   \\ \hline
            M3SNet & 47.989 & 0.939 & 33.147 & 53.98 & 0.926 & 31.04 \\
            M3SNet+ & 49.173 & 0.938 & 33.062 & 49.9 & 0.934 & 32.897 \\
            M3SNet++ & \textbf{46.889} & \textbf{0.943} & \textbf{33.173} & \textbf{48.609} & \textbf{0.94} & \textbf{33.097}  \\
            $\Delta\uparrow$ & 1.100 & 0.004 & 0.026 & 5.371 & 0.014 & 2.057  \\ \hline
            SFNet & 48.807 & 0.939 & 33.164 & 51.351 & 0.932 & 31.802 \\
            SFNet+ & 46.155 & 0.941 & 33.303 & 48.798 & 0.936 & 33.213 \\
            SFNet++ & \textbf{44.982} & \textbf{0.946} & \textbf{33.316} & \textbf{46.852} & \textbf{0.944} & \textbf{33.205} \\
            $\Delta\uparrow$ & 3.825 & 0.007 & 0.152 & 4.499 & 0.012 & 1.403 \\ \hline
            Restormer & 50.785 & 0.932 & 32.821 & 53.598 & 0.929 & 31.446  \\
            Restormer+ & 47.795 & 0.938 & 33.251 & 48.575 & 0.936 & 33.188  \\
            Restormer++ & \textbf{46.476} & \textbf{0.943} & \textbf{33.397} & \textbf{46.57} & \textbf{0.943} & \textbf{33.292} \\
            $\Delta\uparrow$ & 4.309 & 0.011 & 0.576 & 7.028 & 0.014 & 1.846  \\ \hline
    
        \end{tabular}
    }
    \label{tab:sirr_result}
\end{table}

\begin{figure}[t] \centering
    \captionsetup[subfloat]{labelfont=scriptsize,textfont=scriptsize}
    \subfloat[Input]{
        \includegraphics[width=0.084\textwidth]{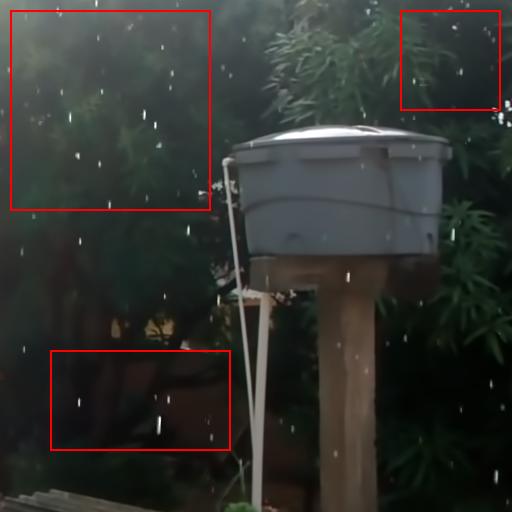}
    }
    \subfloat[PReNet]{
        \includegraphics[width=0.084\textwidth]{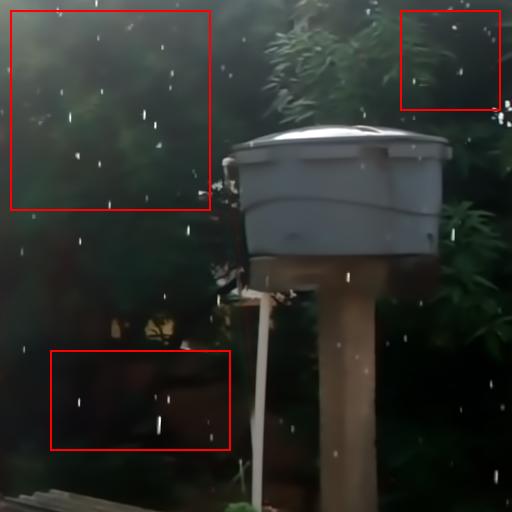}
    }
    \subfloat[M3SNet]{
        \includegraphics[width=0.084\textwidth]{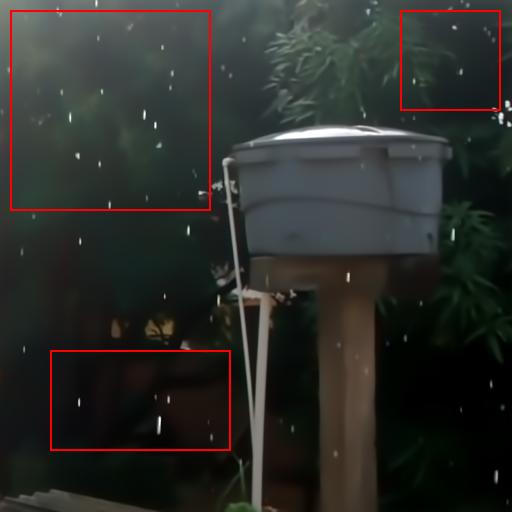}
    }
    \subfloat[SFNet]{
        \includegraphics[width=0.084\textwidth]{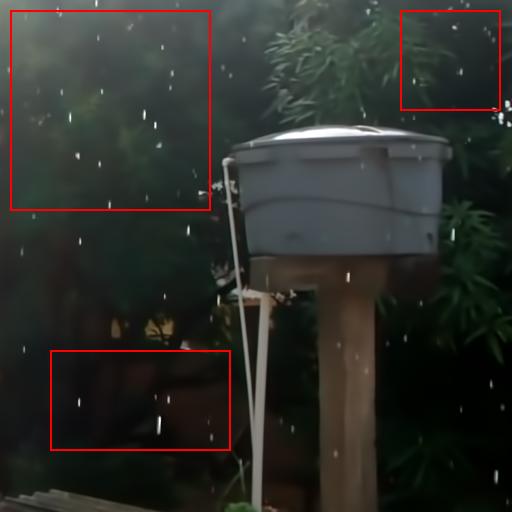}
    }
    \subfloat[Restormer]{
        \includegraphics[width=0.084\textwidth]{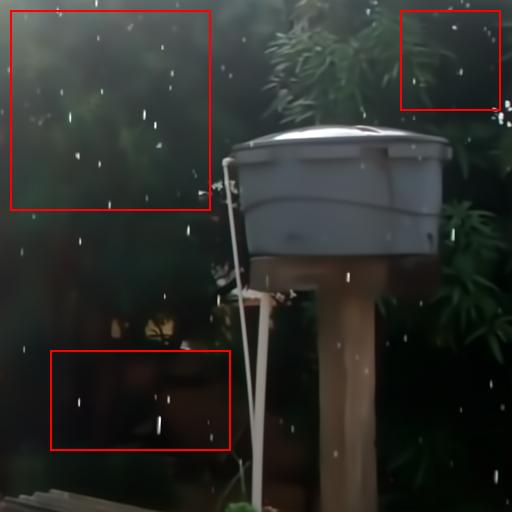}
    }
    \\
    \subfloat[Ground truth]{
        \includegraphics[width=0.084\textwidth]{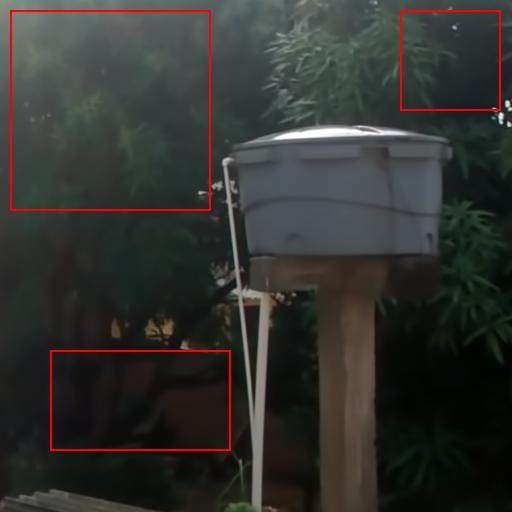}
    }
    \subfloat[PReNet++]{
        \includegraphics[width=0.084\textwidth]{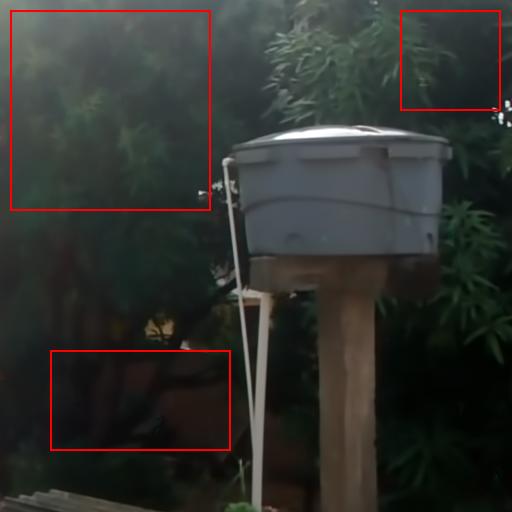}
    }
    \subfloat[M3SNet++]{
        \includegraphics[width=0.084\textwidth]{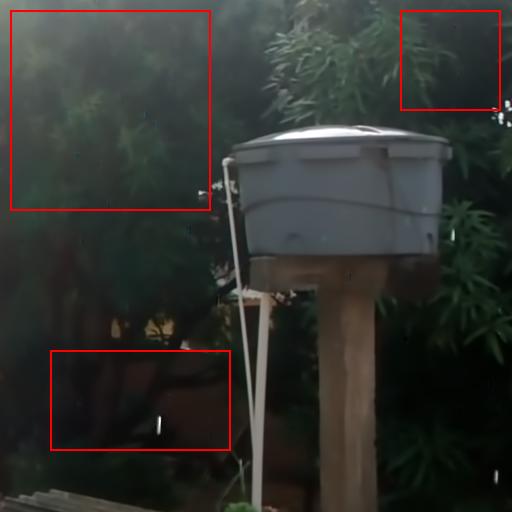}
    }
    \subfloat[SFNet++]{
        \includegraphics[width=0.084\textwidth]{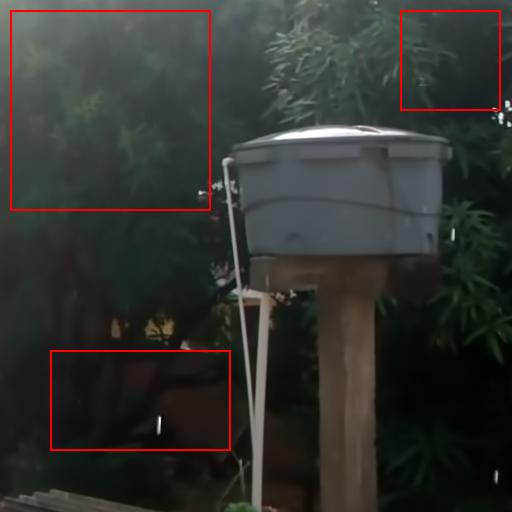}
    }
    \subfloat[Restormer++]{
        \includegraphics[width=0.084\textwidth]{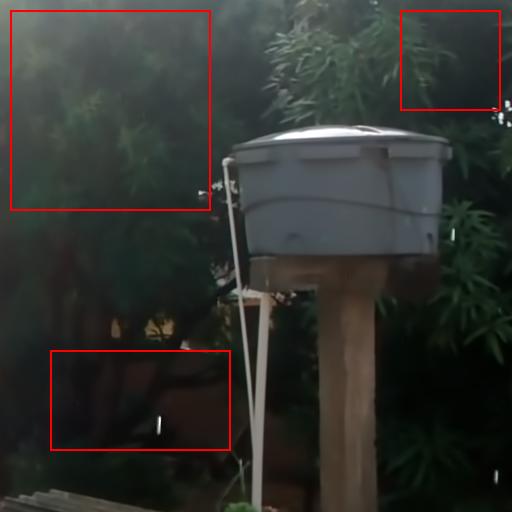}
    }
    \caption{Visual comparison of rain removal results on a test image from SPA-Data. The first row is the input rainy image and the output of derainers trained on the original Rain1400 train set. The second row is the ground truth background image and the output of derainers trained on the Rain1400 train set augmented by our HRIGNet.} 
    \label{fig:figure_sirr}
\end{figure}

\textbf{Image deraining.} With HRIGNet, sufficient rainy images can be conveniently generated from background images and rain layer masks, enabling us to create new paired rainy image datasets or augment existing datasets. In this subsection, we use HRIGNet to augment the existing datasets with ratio 1, further improving the deraining performance of current DL-based derainers on real rain datasets.

We evaluate the effectiveness of the augmentation strategy benefitted from HRIGNet with latest DL-based SIRR methods, including PReNet \cite{ren2019progressive}, M3SNet \cite{gao2023mountain}, SFNet \cite{cui2022selective} and Restormer \cite{zamir2022restormer}. The train sets are common synthetic datasets, including RainTrainL \cite{yang2017deep} and Rain1400 \cite{fu2017removing}. We augment these datasets, retrain the derainers and compare their generalization performance on the real dataset SPA-Data \cite{wang2019spatial}.

The quantitative comparison is shown in Table \ref{tab:sirr_result}, where ``+'' denotes the augmented training with physics-based rendering \cite{halder2019physics} (where we use to produce rain layer masks for our HRIGNet) and ``++'' denotes the augmented training with our method. $\Delta\uparrow$ represents the performance gain brought by the augmented training with our method. As seen, our method improves the performance of all derainers to varying degrees. Fig. \ref{fig:figure_sirr} shows the visual comparison of rain removal results on a test image from SPA-Data. Under such a complex rainy scene, these derainers with augmented training evidently remove rains. The results validate that the rainy images generated by HRIGNet can help improve the accuracy of these deep derainers to rainy images in the real world, thus indirectly validating the realism of the generated rainy images.

\begin{figure}[t] \centering
    \includegraphics[width=0.488\textwidth]{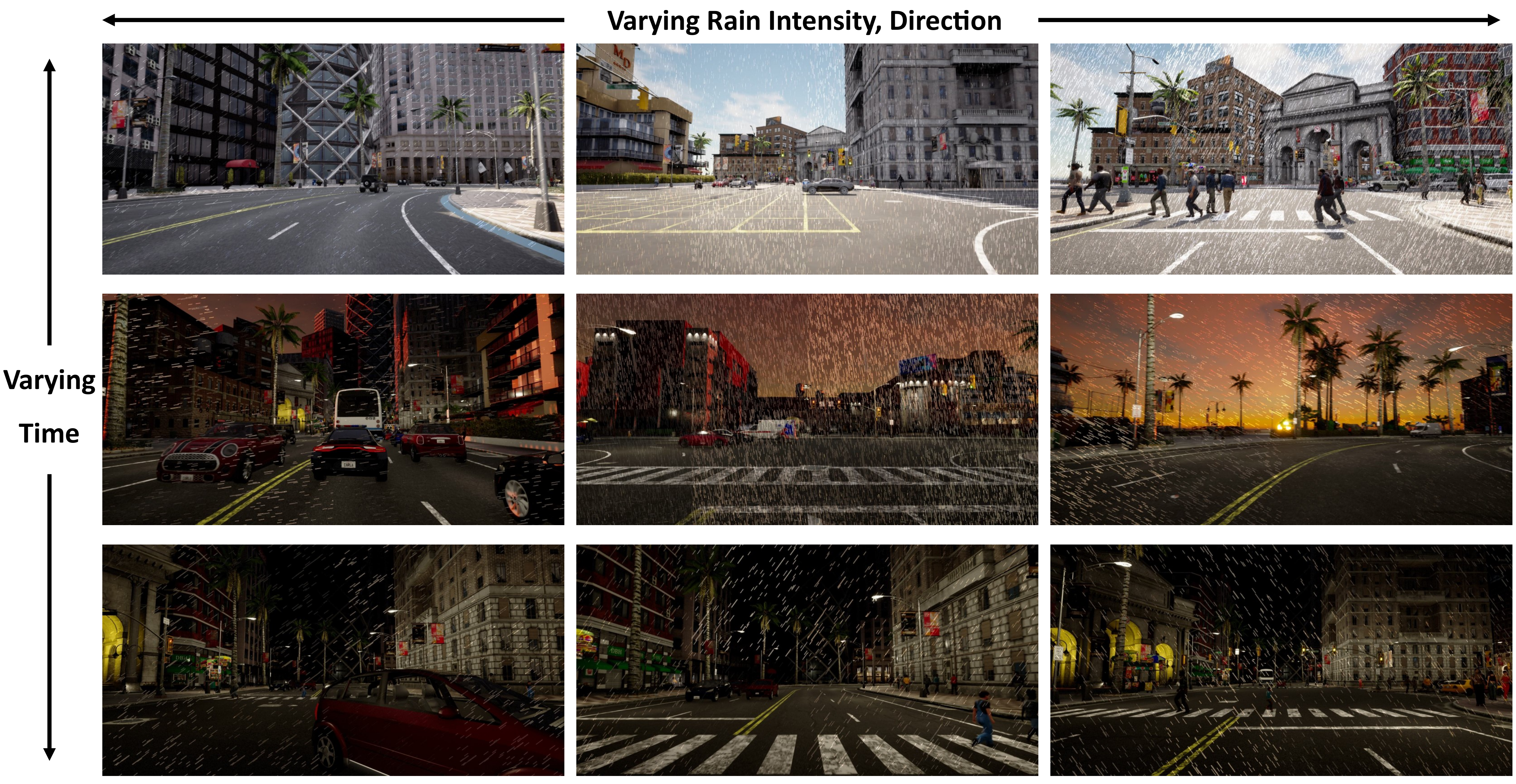}
    \caption{Samples of rainy images from ExtremeRain with multi-attribute controllability (varying in scenes, time, rain intensity and direction). } 
    \label{fig:figure_rain_controlbility}
\end{figure}

\textbf{Controlibility.} As shown in Figure \ref{fig:figure_rain_controlbility}, some rainy images from ExtremeRain are presented. By combining HRIGNet with CRIGNet and CARLA, it is possible to control different background scenes, achieve variations in illumination such as daytime, sunset, and night, and control attributes like rain intensity and direction. The controllability of multiple attributes ensures the diversity of the dataset.

\subsection{Semantic Segmentation In Extreme Rainfall}

\begin{figure}[t] \centering
    \includegraphics[width=0.488\textwidth]{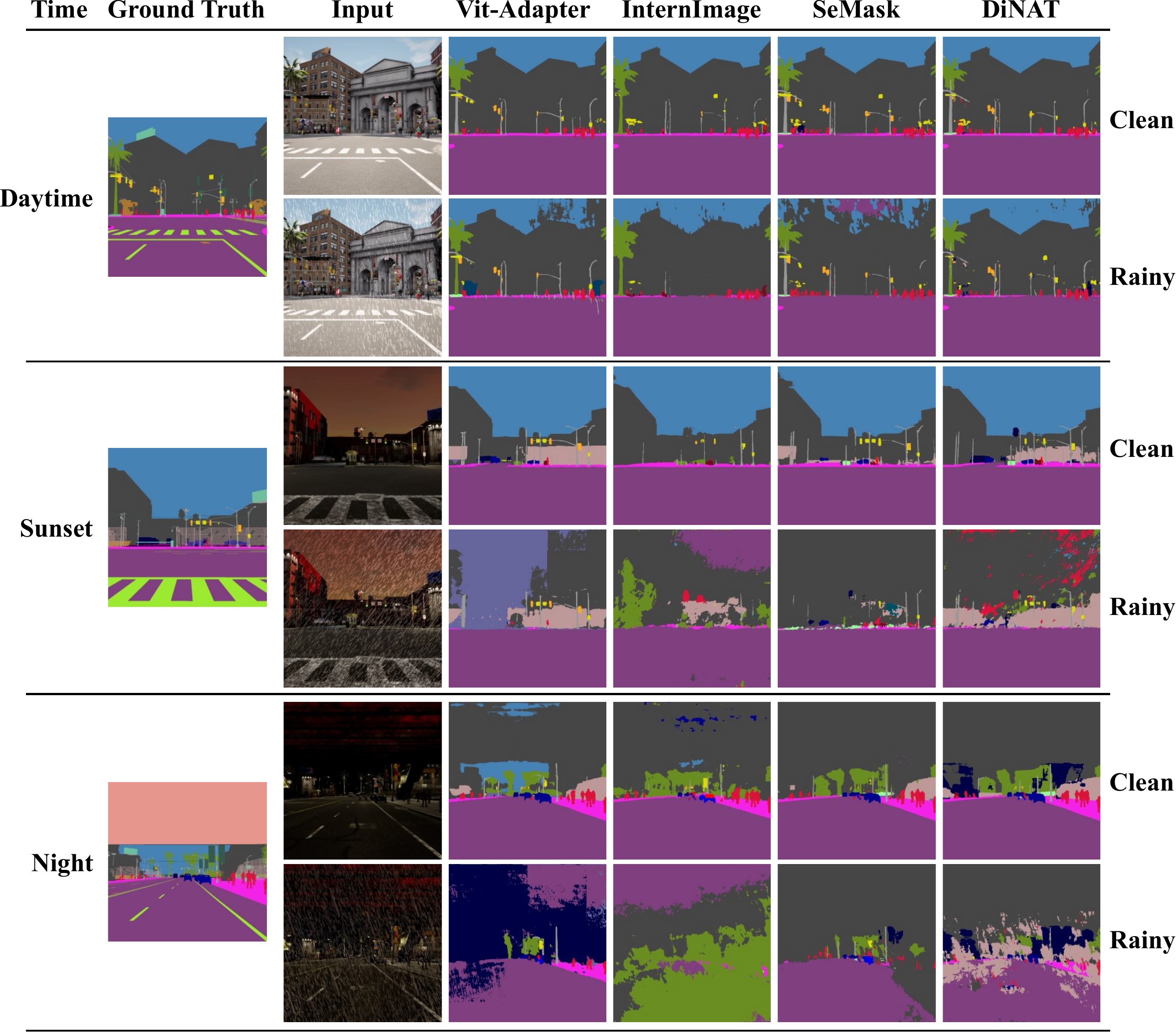}
    \caption{Visual comparisons of semantic segmentation results on clean and rainy RGB images from ExtremeRain dataset.} 
    \label{fig:figure_semantic_evaluation}
\end{figure}

\begin{figure}[t] \centering
    \includegraphics[width=0.488\textwidth]{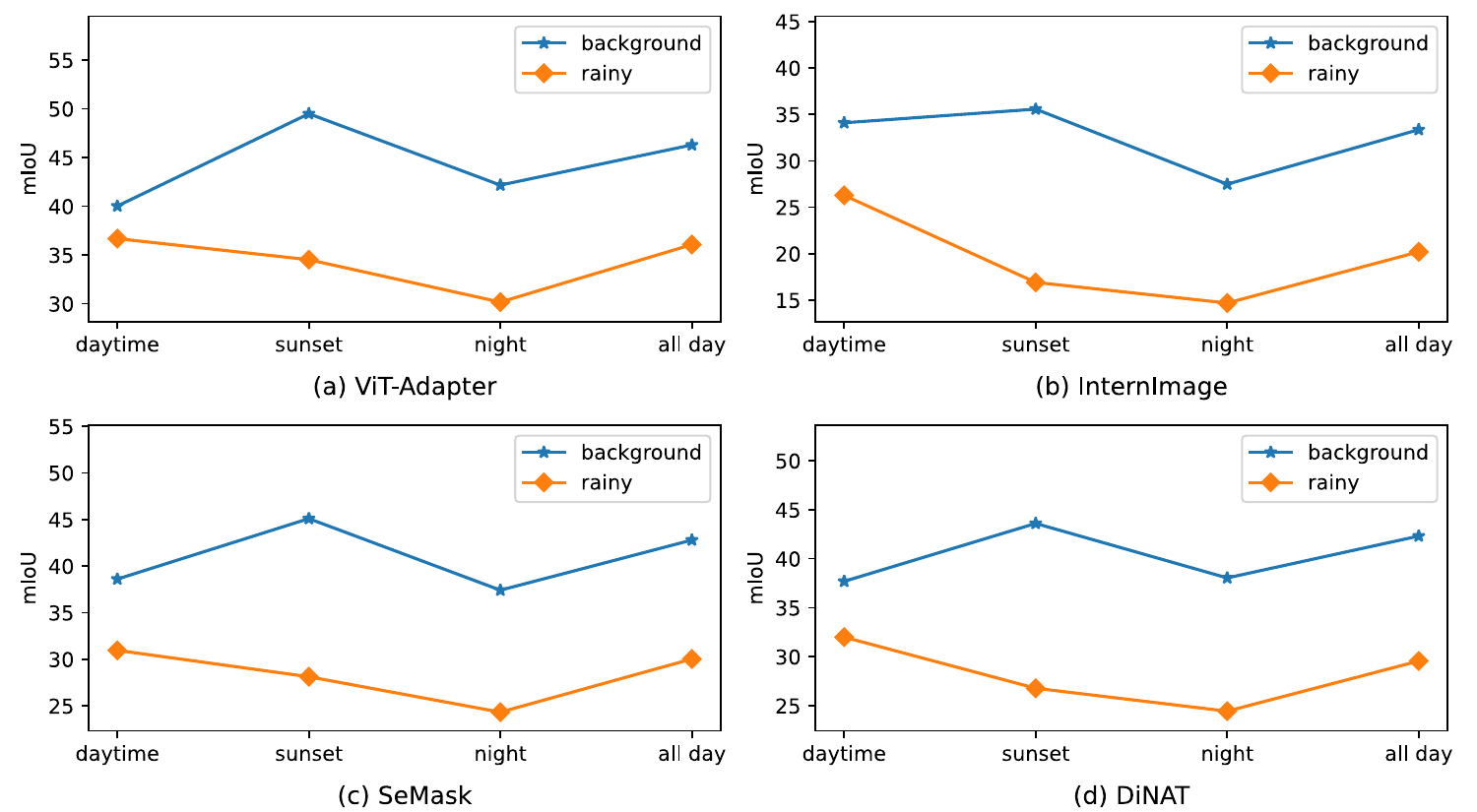}
    \caption{Semantic segmentation quantitative results (mIoU) on clean and rainy RGB images from ExtremeRain dataset.} 
    \label{fig:plot_semantic_evaluation}
\end{figure}

\textbf{Semantic segmentation evaluation.} To evaluate the impact of extreme rainy scenes under different illumination conditions, we test several SOTA semantic segmentation models (including ViT-Adapter\cite{chen2022vision}, InternImage\cite{wang2023internimage}, SeMask\cite{jain2023semask}, and DiNAT\cite{hassani2023neighborhood}) on clean and rainy RGB images from the ExtremeRain dataset respectively. These models are all pretrained on the Cityscapes \cite{cordts2016cityscapes} dataset. Visual comparisons are shown in Figure \ref{fig:figure_semantic_evaluation}, while quantitative results are presented as line charts in Figure \ref{fig:plot_semantic_evaluation}. It can be seen from these results that the accuracy of all models decreases in  rainy scenes. In daytime scenarios, with simple illumination conditions, rainy scenes have a minor impact on model accuracy. By contrast, in sunset and night scenarios, the dual impact of complex illumination and extreme rainfalls leads to a significant drop in semantic segmentation accuracy.


\textbf{Semantic segmentation augmentation.} To improve the accuracy of semantic segmentation models in extreme rainy scenes, we further conduct augmented training with the ExtremeRain dataset and evaluate the models on synthetic and real datasets.

Firstly, We select 5,000 rainy images from discrete domains of the synthetic dataset SHIFT \cite{sun2022shift} as the test set for quantitative evaluation (mIoU). We separately conduct augmented training on semantic segmentation models using background and rainy images from the ExtremeRain dataset, and compare their prediction results, as shown in Table \ref{tab:semantic_evaluation_shift},  where ``Bg-Aug'' denotes augmented training with background images and ``Rainy-Aug'' denotes augmented training with rainy images. $\Delta\uparrow$ represents the performance gain brought by augmented training. The results show that the accuracy of all augmented models improves, and those trained on rainy images show a greater improvement, with their accuracy improved by $5\% - 8\%$. This indicates that the pre-trained model's original dataset lacks diversity, particularly extreme rainy images, while our dataset effectively compensates for this deficiency, thereby enhancing model accuracy.

\begin{figure}[ht] \centering
    \includegraphics[width=0.488\textwidth]{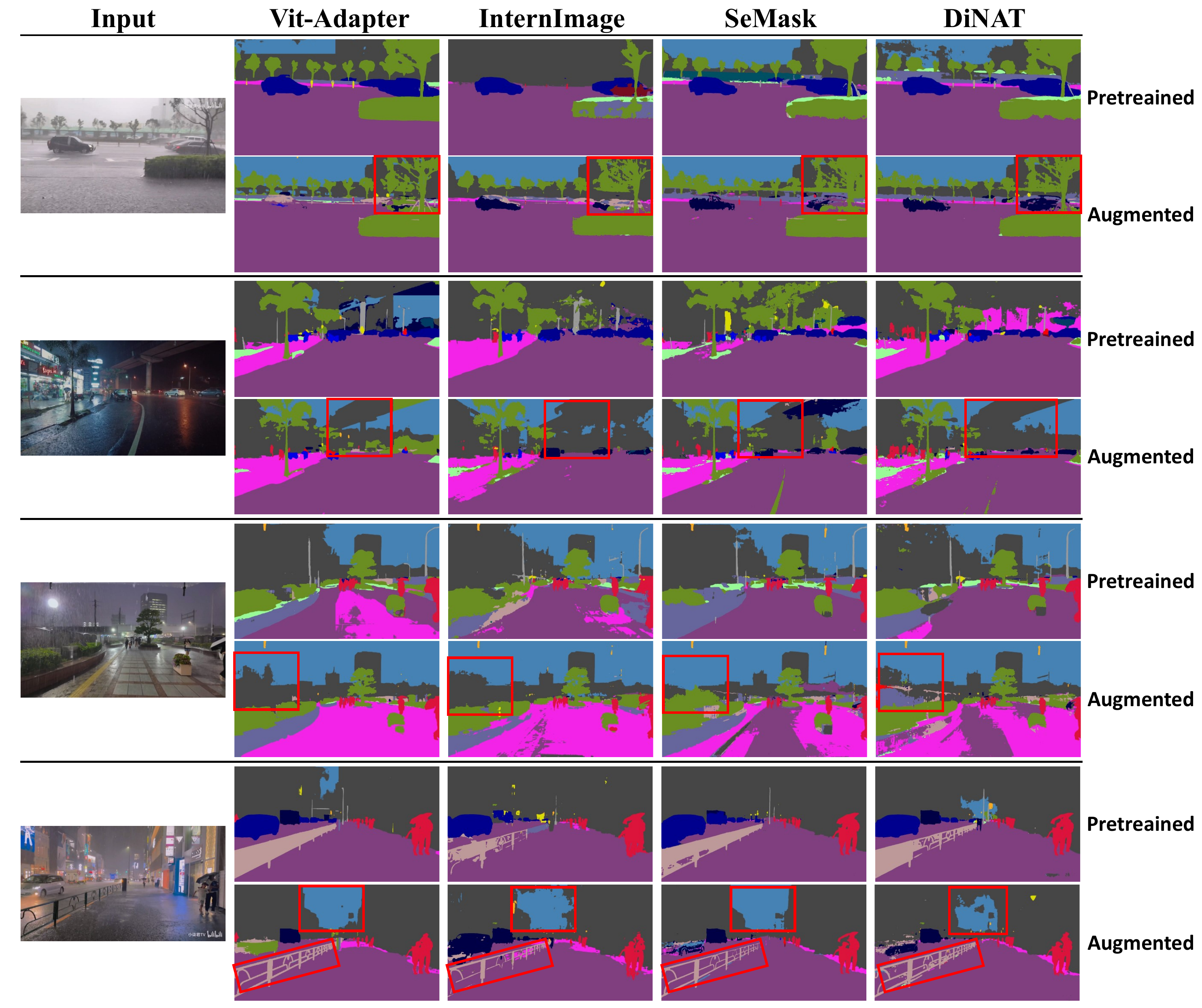}
    \caption{Visual comparisons of semantic segmentation results on real data. ``Pretrained'' denotes prediction results of pre-trained models, and ``Augmented'' denotes those augmented with rainy images from ExtremeRain.} 
    \label{fig:figure_semantic_test_real}
\end{figure}

Finally, we conduct a qualitative evaluation using real extreme rainy street scene images to validate that ExtremeRain dataset can improve performance of semantic segmentation models in real rainy scenes. We collect real rainy scene images with different illumination conditions from the Internet, and use them as the test set. As shown in Figure \ref{fig:figure_semantic_test_real}, we visually compare the prediction results of pre-trained models and those augmented with rainy images on real rainy images. Results indicate that augmented training enables models to better recognize object contours (e.g., buildings and sky) in long-shot views and refine segmentation of nearby objects (e.g., trees and lane lines). In extreme rainy scenarios under complex illuminations, where lighting is poor and rain streaks prominent, the segmentation performance improves significantly.

However, augmented training may cause models to overfit to rainy scene features (e.g., rain streaks), reducing prediction accuracy for objects resembling rain streaks, like utility poles, as the overemphasis on learning rain-specific patterns can blur the distinction between such features and the core visual cues of similar objects, leading to misclassification. Thus, future work should optimize the training strategy. For instance, combining our extreme rainy scene dataset with existing ones to form a new dataset and training models anew can balance their adaptability to both rainy and clean scene features.

\begin{table}\centering
    \caption{Quantitative results (mIoU) on the SHIFT dataset.}
    \begin{tabular}{c|c|cc|cc}
        \hline
        \multirow{2}{*}{target} & Pretrained & \multicolumn{2}{c|}{Bg-Aug} & \multicolumn{2}{c}{Rainy-Aug} \\
        ~ & predict & predict & $\Delta\uparrow$ & predict & $\Delta\uparrow$  \\
        \hline
        Vit-Adapter & 37.38 & 39.55 & 2.17 & \textbf{42.89} & \textbf{5.51} \\
        InternImage & 30.88 & 34.79 & 3.91 & \textbf{38.12} & \textbf{7.24} \\
        SeMask & 32.28 & 38.80 & 6.52 & \textbf{40.26} & \textbf{7.98} \\
        DiNAT & 34.86 & 38.14 & 3.28 & \textbf{40.58} & \textbf{5.72} \\
        \hline
    \end{tabular}
    \label{tab:semantic_evaluation_shift}
\end{table}

\section{Conclusion And Limitations}
\label{section:conclusion}

Existing rainy image synthesizers often suffer from poor illumination controllability and limited realism, which significantly undermines the effectiveness of the model evaluation. To that end, we propose a learning-from-rendering rainy image synthesizer, which combines the benefits of realism and controllability. By integrating the proposed synthesizer with the CARLA driving simulator, we develop CARLARain—an extreme rainy street scene simulator which can obtain paired rainy-clean images and labels under complex illumination conditions. Qualitative and quantitative experiments validate that CARLARain can effectively improve the accuracy of semantic segmentation models in extreme rainy scenes.

However, there are still avenues for further improvements. Our rainy image synthesizer combines the two stages via rainy image datasets, limiting early dataset creation efficiency. In future work, we would like to integrate them more effectively. Additionally, establishing a refined feedback mechanism between CARLARain and autonomous driving systems—enabling the simulator to receive system decisions for joint optimization—merits future exploration.

\appendices

\bibliographystyle{IEEEtran}
\bibliography{refs}

\title{Supplementary Material}

\author{Kaibin Zhou, Kaifeng Huang,~\IEEEmembership{Member,~ACM}, Hao Deng,~\IEEEmembership{Member,~IEEE}, Zelin Tao,\\ Ziniu Liu, Lin Zhang,~\IEEEmembership{Senior Member,~IEEE}, and Shengjie Zhao,~\IEEEmembership{Senior Member,~IEEE}
}


\maketitle

\section{Raindrop Particle Physics Simulator}
\label{append:simulation}

In the rendering stage of our proposed rainy image synthesizer, to match real-world raindrop size and distribution, we implement a raindrop particle physics simulator. Next, we will introduce detail derivations about the simulator.

In our simulator, rain attributes like intensity and direction are controllable input parameters. Rain intensity, defined as the water volume per unit area per unit time \cite{duhanyan2011below}, influences raindrop size distribution (RSD) and terminal velocity. We adopt Marshall and Palmer's exponential RSD \cite{best1950size} and Kessler's terminal velocity \cite{kessler1969distribution}. For simplicity, we use a random uniform distribution for initial raindrop spatial distribution and adjust raindrop velocity according to input wind parameters.

In atmospheric modelling, scientists usually quantify rain using a parameter called intensity of rain, $I$. It is the volume of water  \cite{duhanyan2011below} delivered to the ground per unit of ground surface and per unit of time: 
\begin{equation}
    I=\frac{\textit{Volume of water}}{\textit{Surface of precipitation}\times \textit{duration}}.
\end{equation}

In the international system of units, the intensity of rain $I$ is in $m\cdot s^{-1}$. Now, in the literature, the intensity of rain is often expressed in $mm\cdot h^{-1}$. The conversion between two units can be expressed as: 
\begin{equation}
    I = x(mm/h) = x\times\frac{10^{-3}}{3600}(m/s).
\end{equation}

\textbf{Raindrop size distributions.} Let $D$ denote the diameter of raindrops. The number concentration of raindrops with a diameter between $D$ and $D+dD$ is then:
\begin{equation}
    dC(D)=N(D)dD,
\end{equation}
where, $N(D)$ is the raindrop size distribution (RSD). In the meteorological studies, exponential functions are often used to fit the observed RSDs.
\begin{equation}
    N(D)=Ae^{-\beta D}
\end{equation}

The exponential RSD of Marshall and Palmer \cite{best1950size} is one of the simplest and the most often used parameterisation to fit the RSDs: 
\begin{equation}
N(D)=8\times 10^6 e^{-4100I^{-0.21}D}.
\end{equation}

In order to sample raindrops based on the RSD, the cumulative distribution function of raindrop diameters is needed. Limiting the range of raindrop diameters to $D\in [D_{min},D_{max}]$, we can get the total number of raindrops in this range as:

\begin{equation}
\begin{aligned}
    N_{total} &=\int_{D_{min}}^{D_{max}} N(D)dD \\&=\frac{A}{\beta}(e^{-\beta D_{min}}-e^{-\beta D_{max}}).
\end{aligned}
\end{equation}

So the probability density function of $D$ is:
\begin{equation}
    P(D)=\frac{N(D)}{N_{total}}.
\end{equation}

The cumulative distribution function of $D$ is:
\begin{equation}
\begin{aligned}
    F(D)&=P(x\le D)=\int_{D_{min}}^{D}P(x)dx\\&=\frac{\int_{D_{min}}^{D}N(x)dx}{N_{total}}=\frac{e^{-\beta D_{min}}-e^{-\beta D}}{e^{-\beta D_{min}}-e^{-\beta D_{max}}}.
\end{aligned}
\end{equation}

In the implementation, based on the variable $u\in [0,1]$ obtained from sampling from a uniform random distribution, $D$ with distribution of $N(D)$ can be obtained by inverse transforming it using $F$.

Let $u=F(D)\in[0,1]$, then\par
\begin{small}
\begin{equation}
    D=F^{-1}(u)=\frac{ln[e^{-\beta D_{min}}-u(e^{-\beta D_{min}}-e^{-\beta D_{max}})]}{-\beta}.
\end{equation} 
\end{small}

\textbf{Raindrop terminal velocity.}  It is generally assumed that the raindrops fall at their terminal velocity, whatever their position (their height) in the atmosphere. The transient period to reach that speed is then totally neglected. We use the raindrops terminal velocity given by Kessler \cite{kessler1969distribution}:
\begin{equation}
    v_t=130 D^{0.5}
\end{equation}



\section{ControlNet-based up-scale model for CRIGNet}

\begin{figure}[ht] \centering
    \includegraphics[width=0.48\textwidth]{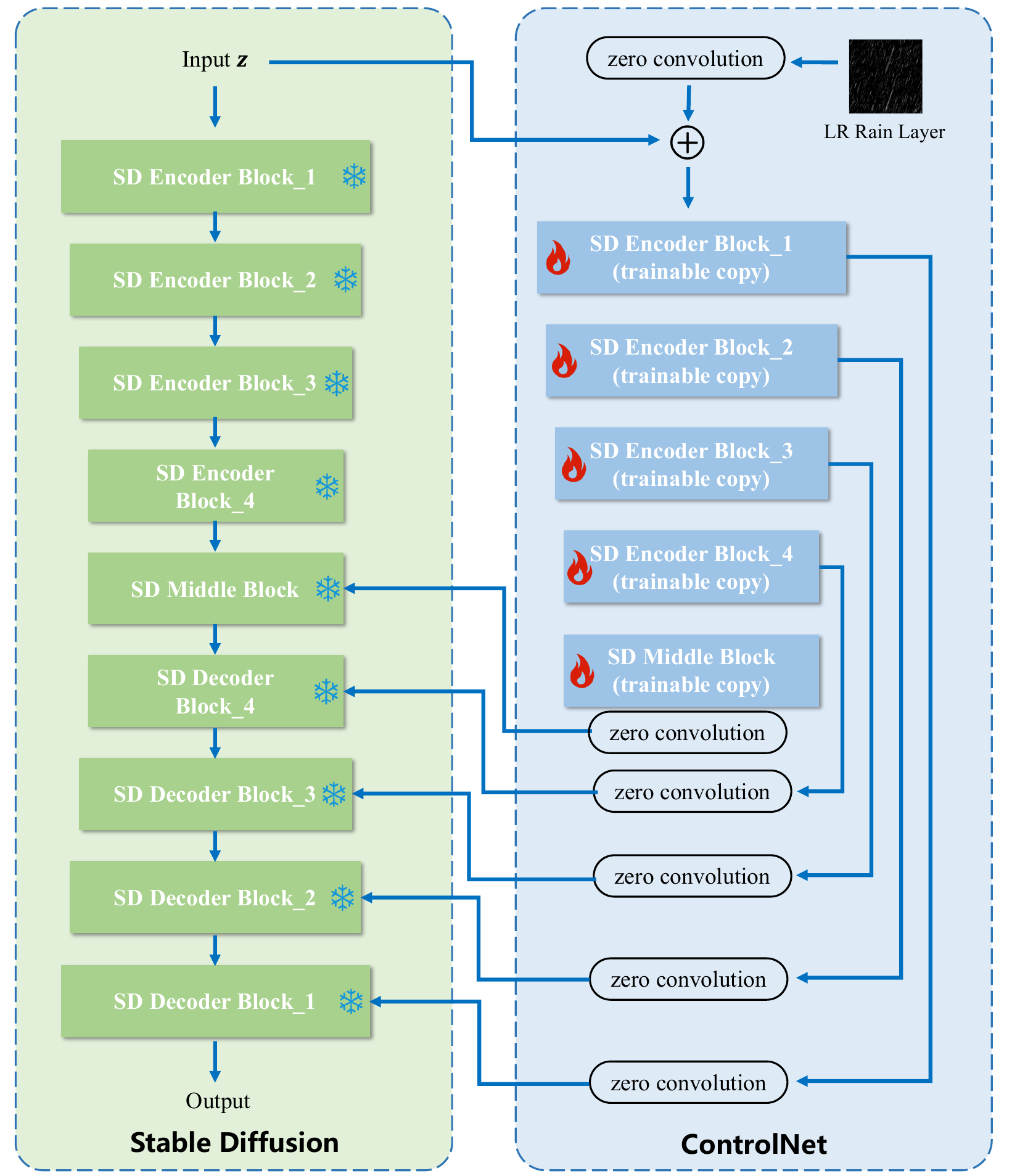}
    \caption{Overview of the ControlNet-based rain streak up-scale model for CRIGNet.} 
    \label{fig:figure_controlnet}
\end{figure}

In the rain streak generation of CARLARain, due to the limitations of the Variational Autoencoder (VAE) on which CRIGNet is built, CRIGNet can only generate rain streak images with a maximum resolution of $256 \times 256$. As shown in Fig. \ref{fig:figure_controlnet}, to generate high-resolution rain streak images, we pretrain a ControlNet-based \cite{zhang2023adding} up-scale model.
Using low-resolution rain streak images from CRIGNet as conditionings, the up-scale model generates corresponding high-resolution ones.

ControlNet introduces a plug-and-play module to the Stable Diffusion model \cite{rombach2022high}. This module guides the image generation process of the diffusion model by adding additional control signals. In this paper, given a high-resolution rain streak image $x_{HR}$, we randomly sample a region $m_{LR}$ from its low-resolution counterpart $x_{LR}$. Pixels outside this region are set to black, resulting in the control condition image $x_{control}$, defined as $x_{control} = x_{HR} \cdot m_{LR}$. Using the pre-trained image encoder $\mathcal{E}$ in Stable Diffusion, we encode the control condition image to obtain $c$, i.e., $c = \mathcal{E}(x_{control})$. In ControlNet, a trainable replica $\Theta_c$ is derived from the original network block $\Theta$ of the U-Net in the diffusion model. During training, $\Theta$ is frozen while $\Theta_c$ is fine-tuned; meanwhile, the control condition $c$ is injected into the diffusion model. Specifically, for each network block:
\begin{equation}
    y_c = F(x; \Theta) + Z\left(F\left(x + Z(c; \Theta_{z1}); \Theta_c\right); \Theta_{z2}\right),
\end{equation}
where $x$ is the input feature of the network block, $y_c$ is the output feature of the network block, $F$ is the forward operation of the network block, $Z$ is the zero convolution layer, and $\{\Theta_{z1}, \Theta_{z2}\}$ are the parameters of the zero convolution layer.

In the diffusion model, given a high-resolution rain streak image $x_{HR}$, its latent variable $z_{HR}$ is obtained by encoding through the image encoder $\mathcal{E}$. Noise is gradually added to $z_{HR}$ to get $z_t$, where $t$ is the time step of adding noise. A network $\epsilon_{\theta}$ is trained to predict the noise in $z_t$, and its loss function is as follows:
\begin{equation}
\mathcal{L} = \mathbb{E}_{z_{\text{HR}}, t, c, \epsilon \sim \mathcal{N}(0, \mathbf{I})} \left[ \left\| \epsilon - \epsilon_{\theta}(z_t, t, c) \right\|_2^2 \right].
\label{equ:loss_control}
\end{equation}

In this paper, we construct a $512\times512$ high-resolution rain streak image dataset using the method in \cite{halder2019physics} to train our rain streak image up-scale model. During training, we randomly crop each high-resolution sample into a $256\times256$ low-resolution image as the control condition, while the original high-resolution image serves as the ground truth. The model is optimized using Eq. \ref{equ:loss_control}. By combining the pre-trained CRIGNet generator with our ControlNet-based up-scale model, we can generate diverse high-resolution rain streak images with specified attributes (e.g., intensity and direction).

\section{Datasets}

\begin{figure*}[t] \centering
    \includegraphics[width=0.32\textwidth]{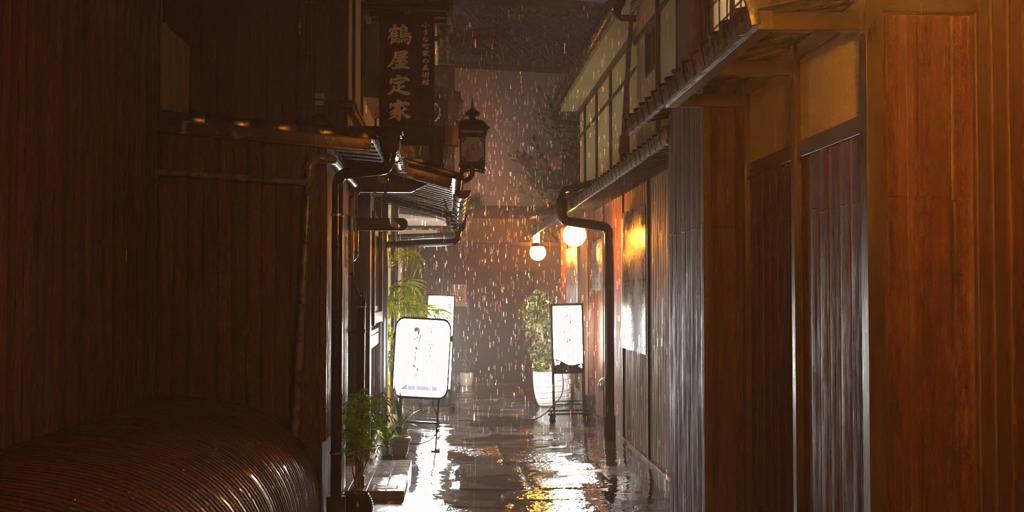}
    \includegraphics[width=0.32\textwidth]{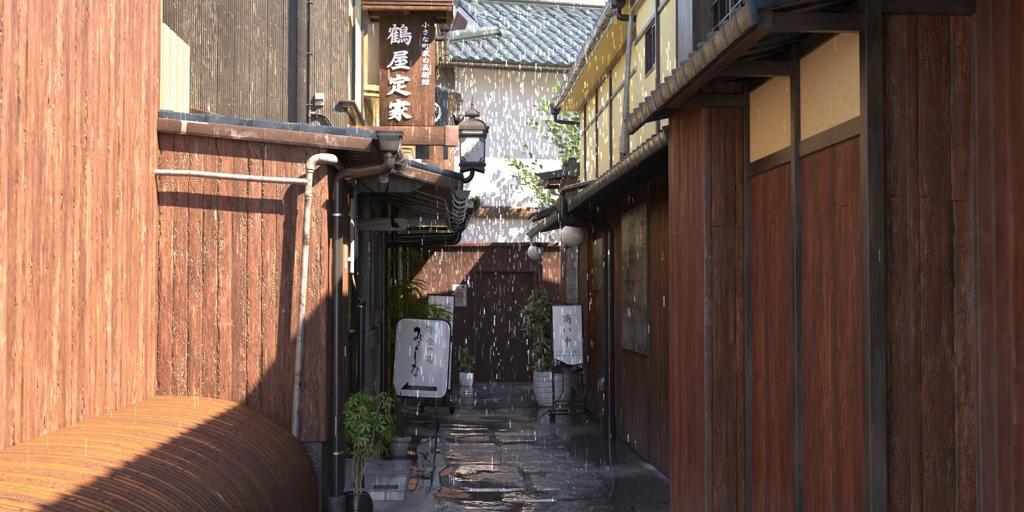}
    \includegraphics[width=0.32\textwidth]{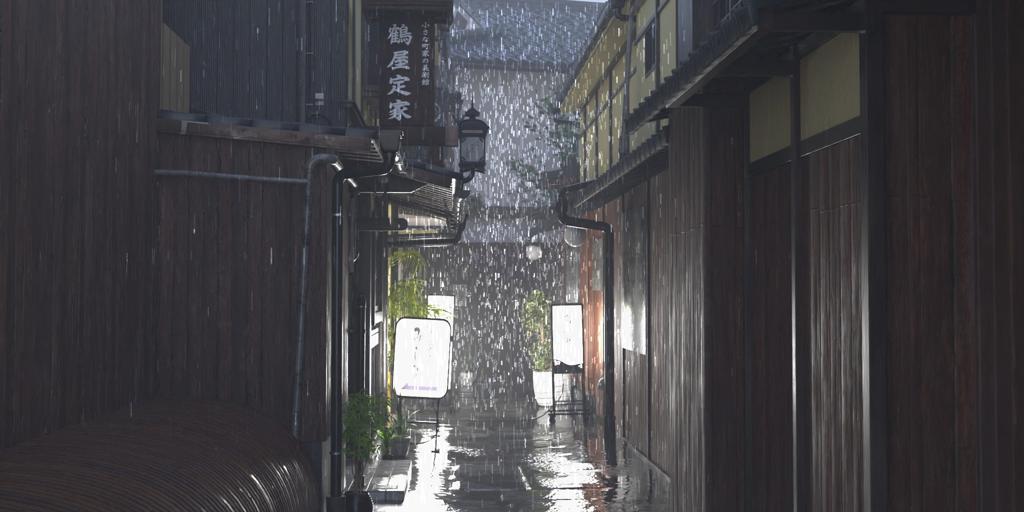}
    \\
    \vspace{0.1cm}
    \includegraphics[width=0.32\textwidth]{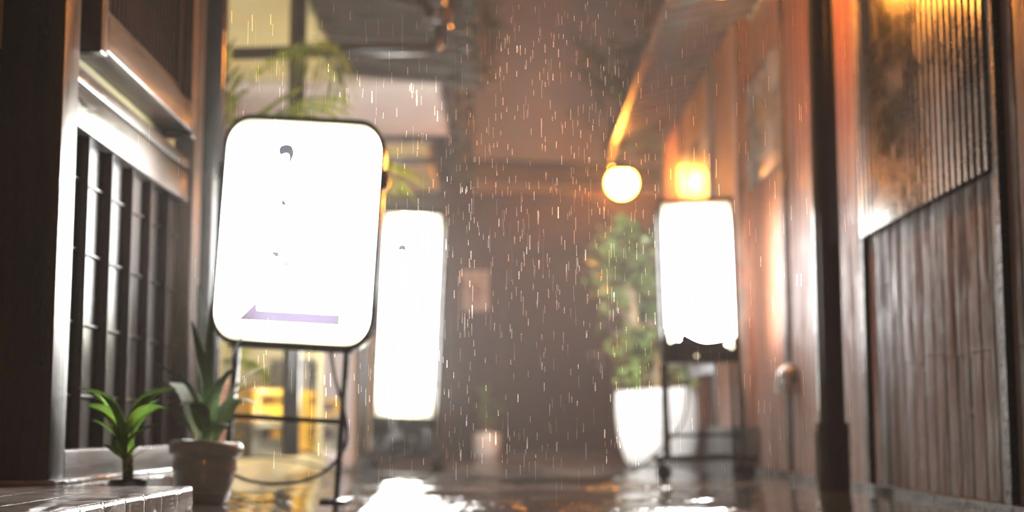}
    \includegraphics[width=0.32\textwidth]{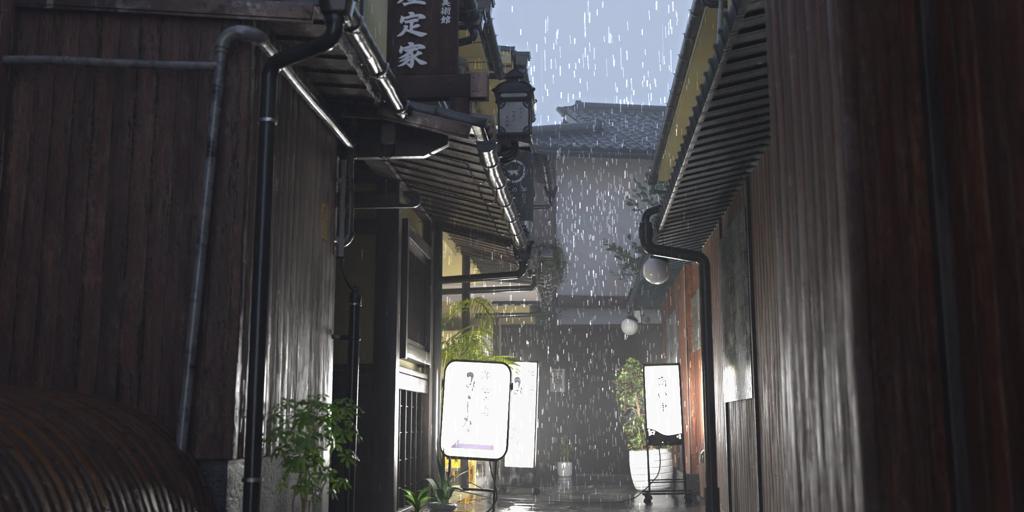}
    \includegraphics[width=0.32\textwidth]{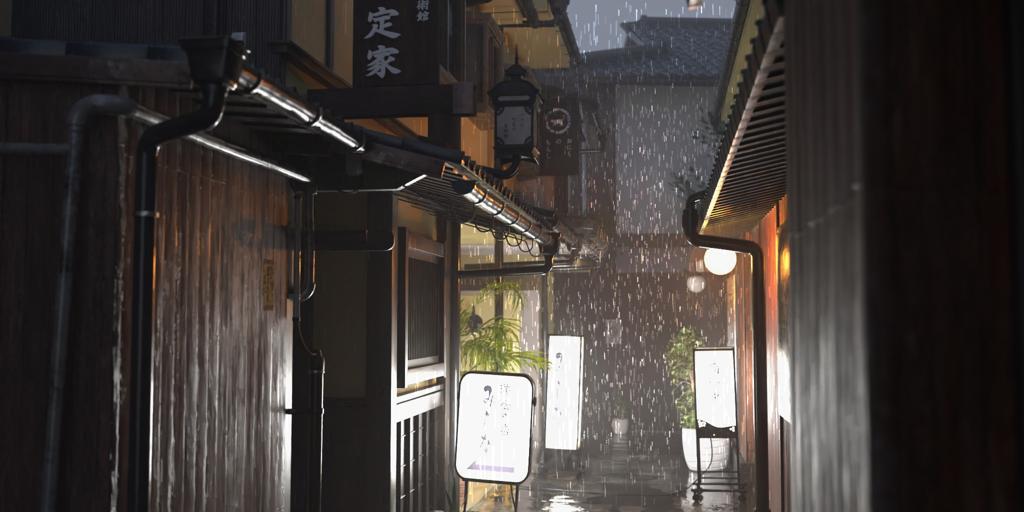}
    \\
    \vspace{0.1cm}
    \includegraphics[width=0.32\textwidth]{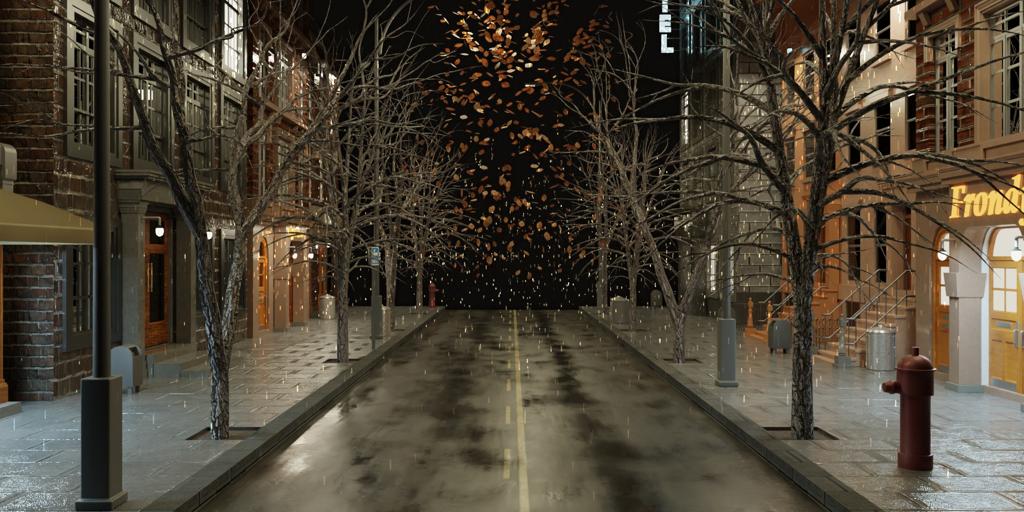}
    \includegraphics[width=0.32\textwidth]{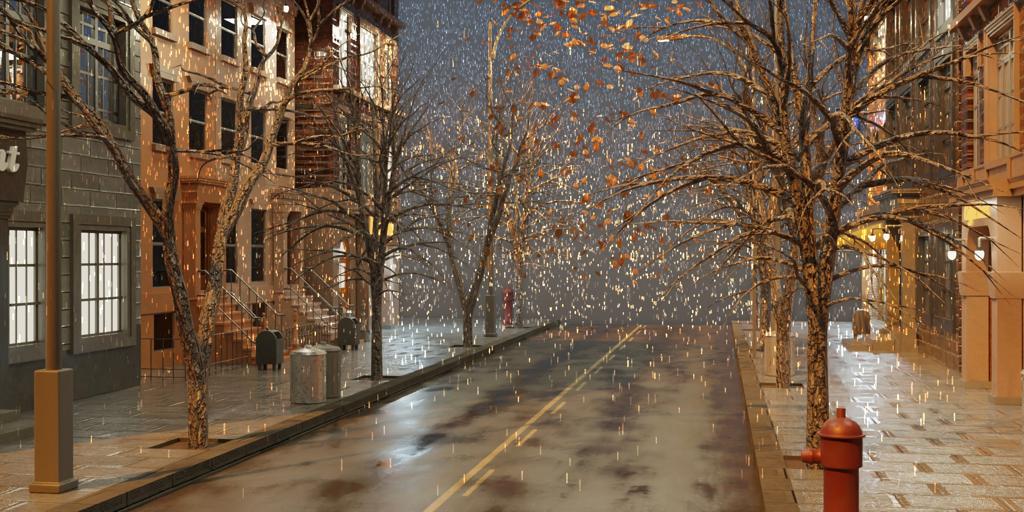}
    \includegraphics[width=0.32\textwidth]{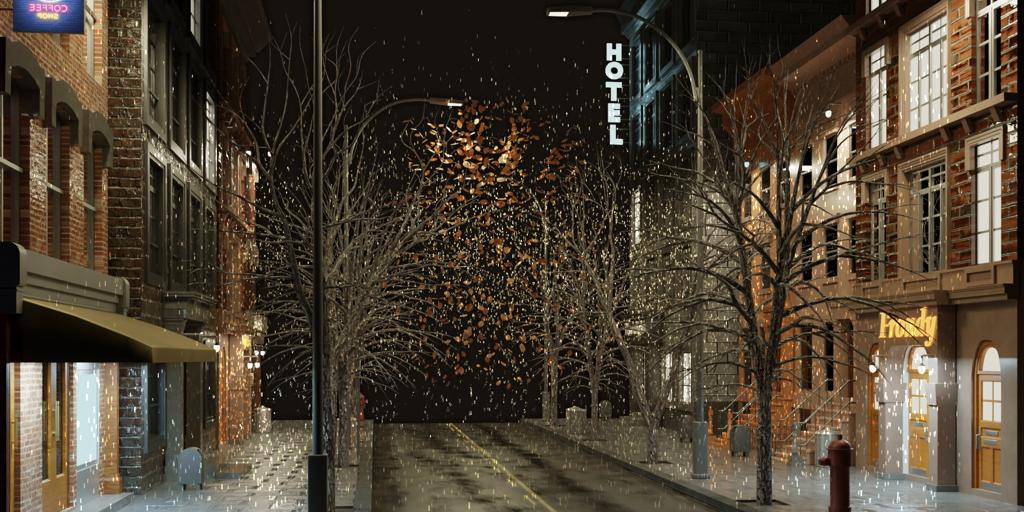}
    \\
    \vspace{0.1cm}
    \includegraphics[width=0.32\textwidth]{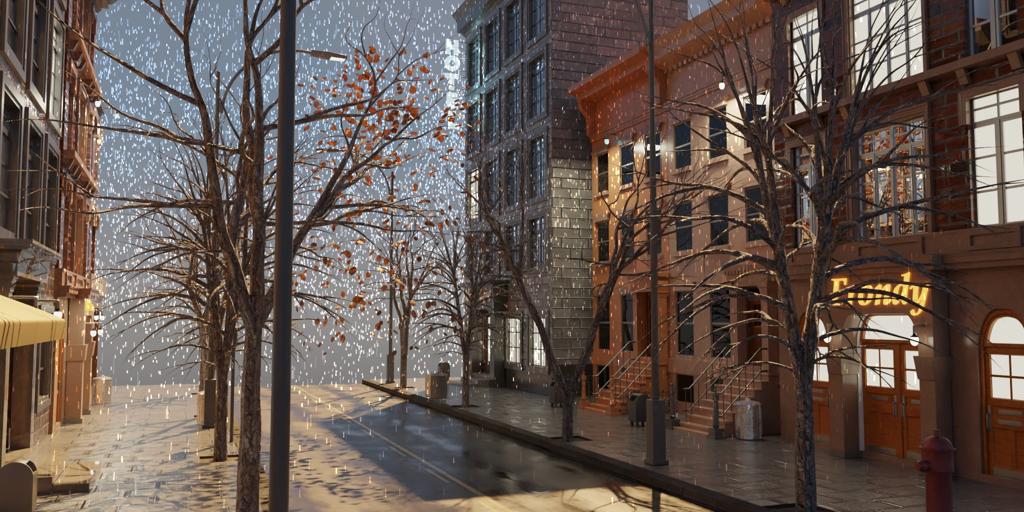}
    \includegraphics[width=0.32\textwidth]{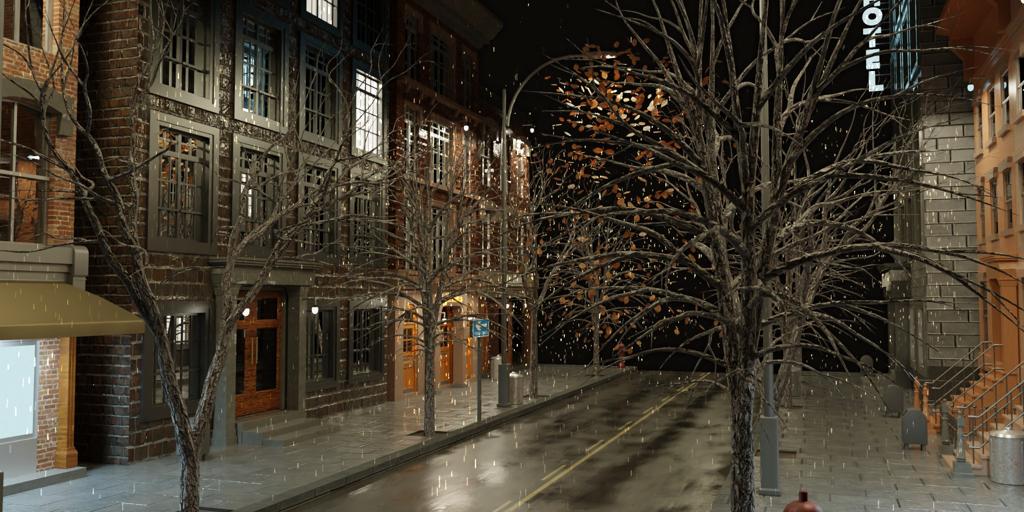}
    \includegraphics[width=0.32\textwidth]{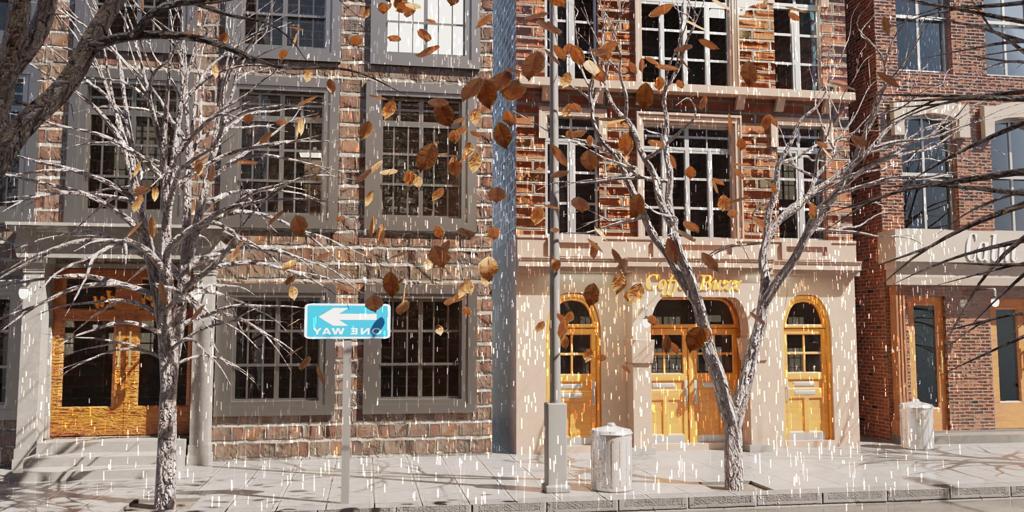}
    \\
    \vspace{0.1cm}
    \includegraphics[width=0.32\textwidth]{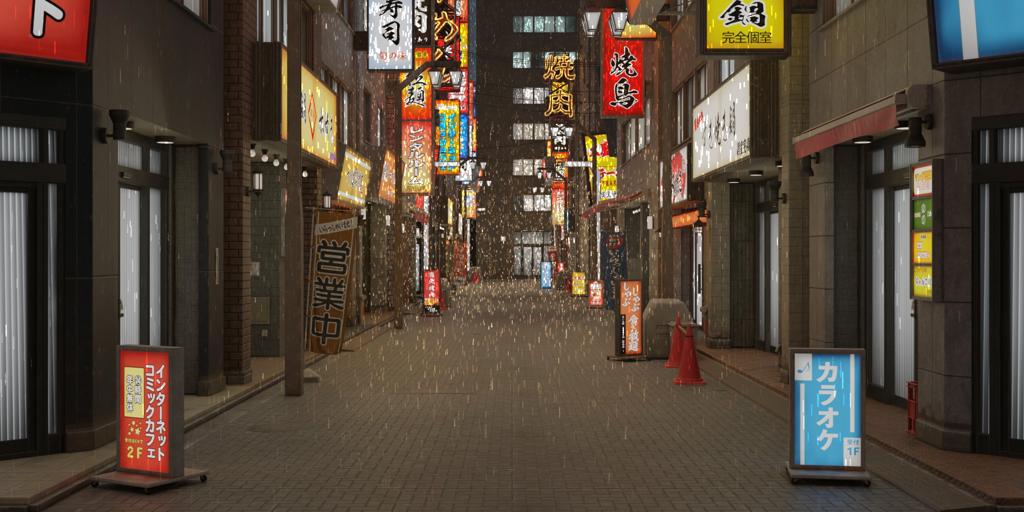}
    \includegraphics[width=0.32\textwidth]{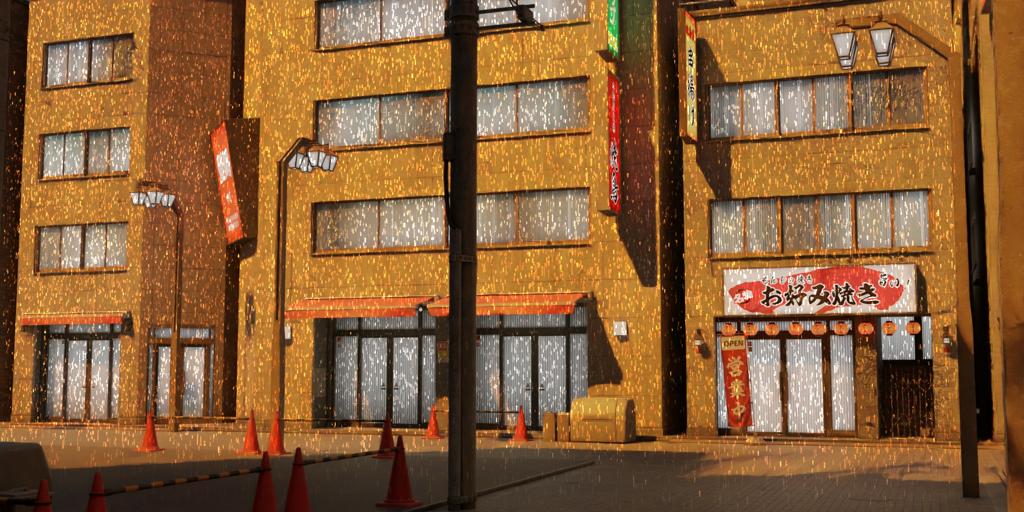}
    \includegraphics[width=0.32\textwidth]{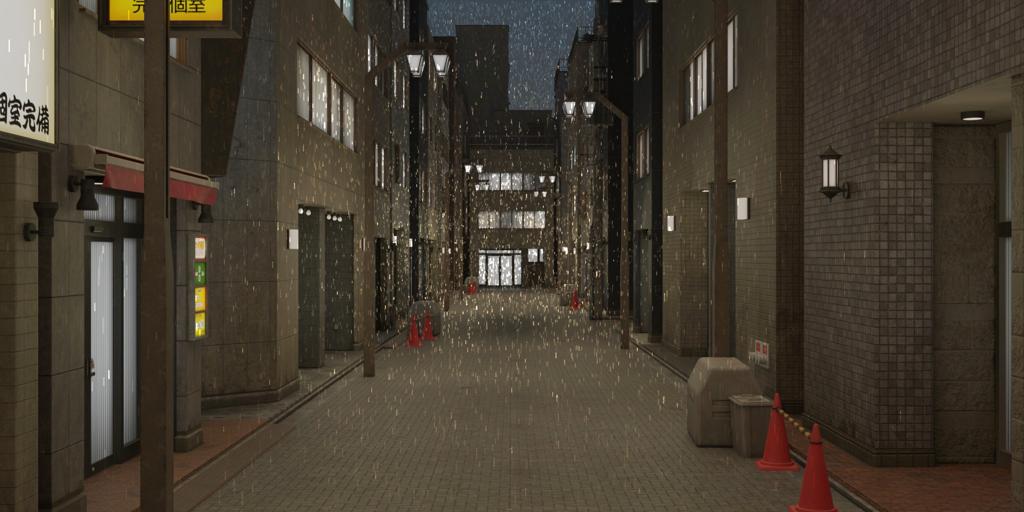}
    \\
    
    \vspace{0.1cm}
    \includegraphics[width=0.32\textwidth]{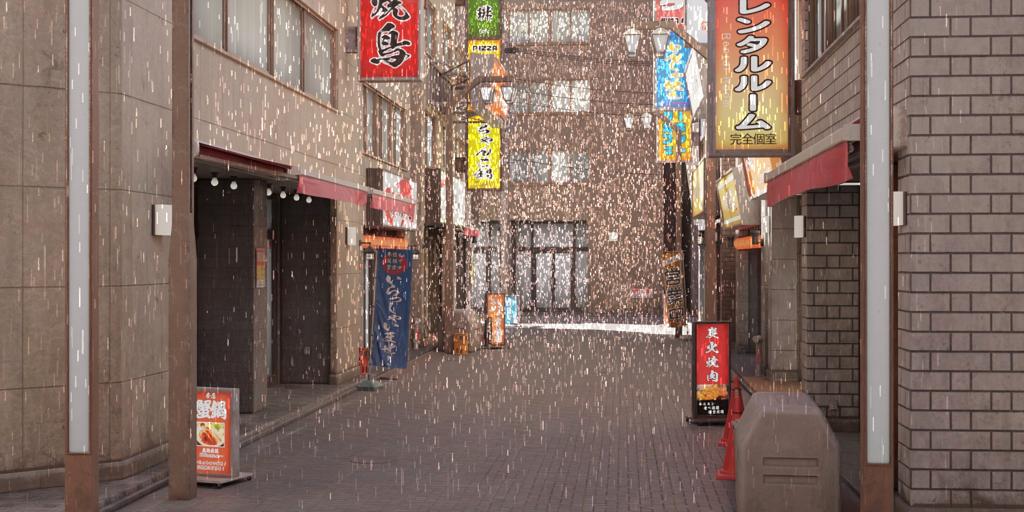}
    \includegraphics[width=0.32\textwidth]{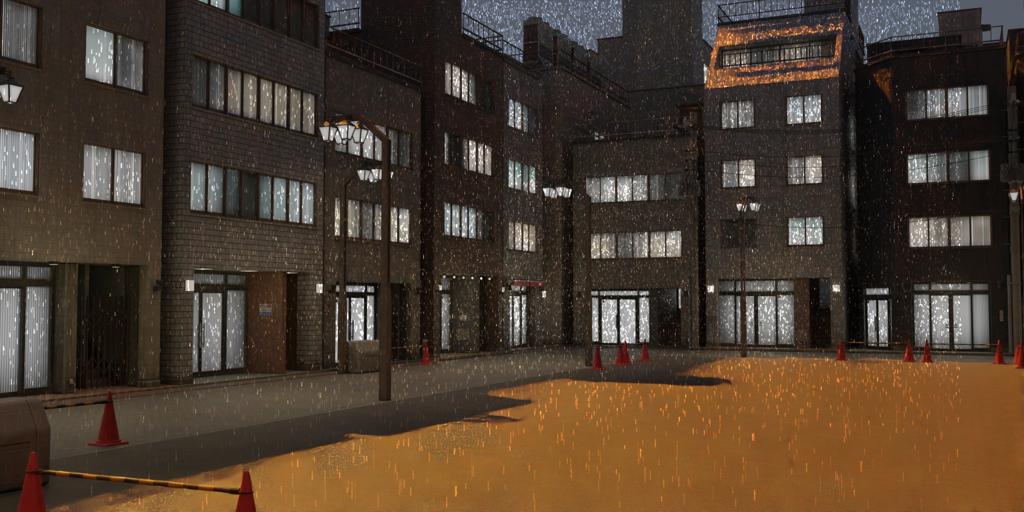}
    \includegraphics[width=0.32\textwidth]{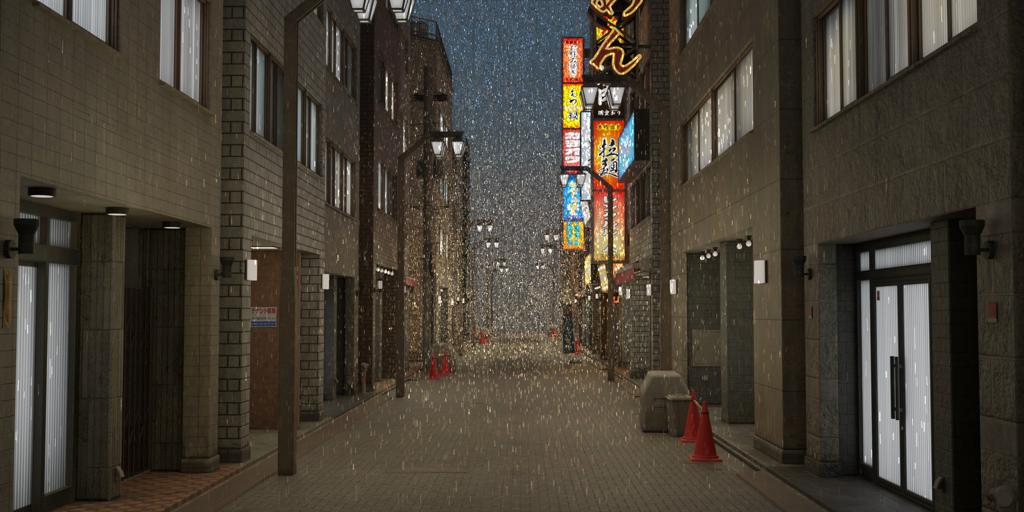}
    \caption{Rainy images of the HRI dataset. The first and second rows are the rainy images of the lane scene. The third and fourth rows are the rainy images of the citystreet scene. The fifth and sixth rows are the rainy images of the japanesestreet scene.} 
    \label{fig:figure_hri}
\end{figure*}

\textbf{High-resolution Rainy Image dataset.} Some rainy images of the High-resolution Rainy Image (HRI) dataset are shown in Fig. \ref{fig:figure_hri}.

\begin{figure*}[h!] \centering
    \includegraphics[width=0.99\textwidth]{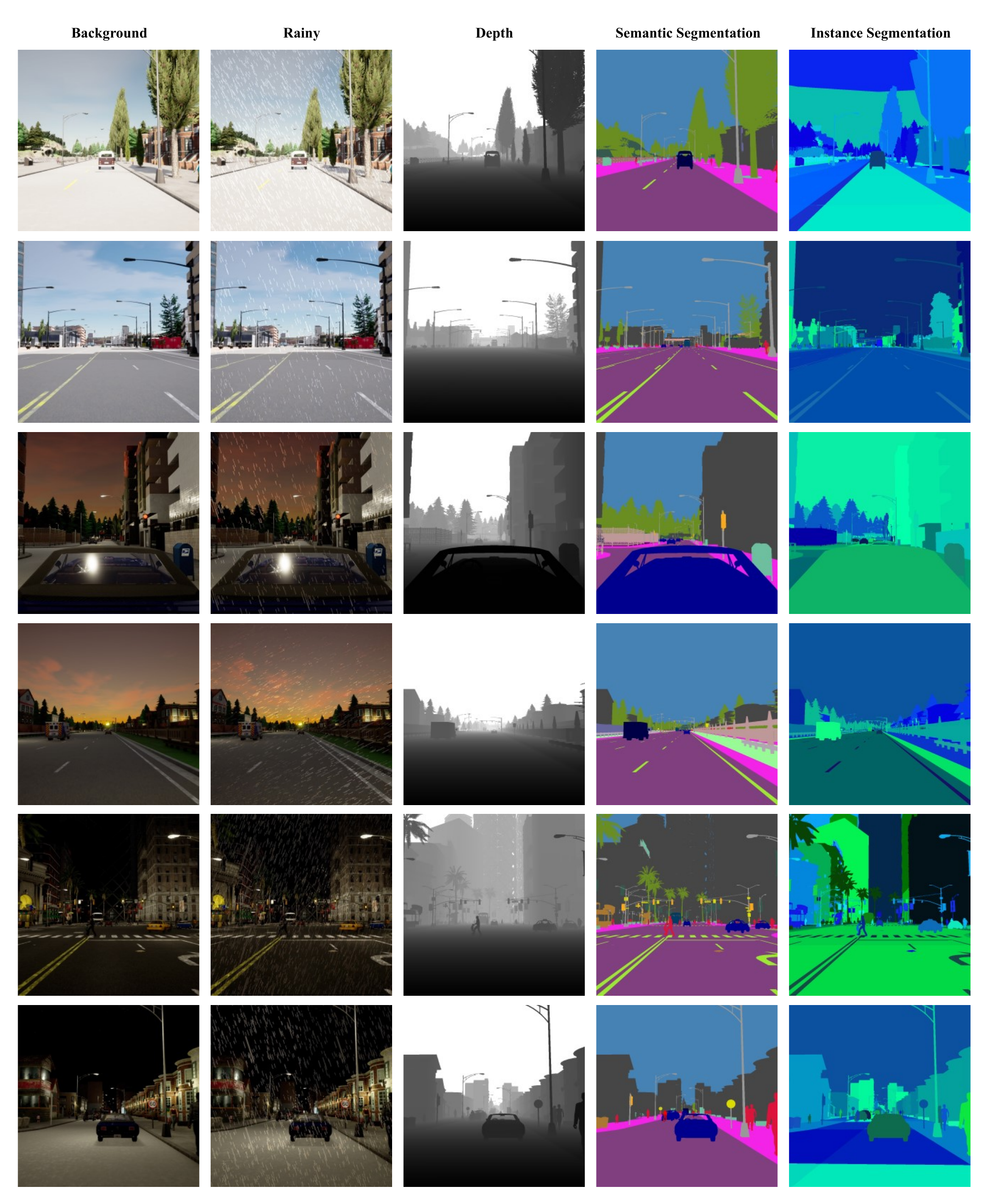}
    \caption{Samples of the ExtremeRain dataset.} 
    \label{fig:figure_extremerain}
\end{figure*}

\textbf{ExtremeRain Dataset.} Some samples of the ExtremeRain dataset are shown in Fig. \ref{fig:figure_extremerain}.

\section{Implementation Details About HRIGNet}
\label{append:experiments}

\subsection{Baselines}
\label{append:experiments_baseline}

For the LDM model in Table V of the main text, the initial learning rate is set as $2\times10^{-6}$, the batch size is 1, the image size of the UNet backbone is $128\times 128$, and the model channels are 224. The first stage model and the conditioning stage model share a same VQGAN model, with the parameters of the model using pretrained vq-f4 from LDM. The training epoch of the LDM model is 99.

For the DiT model in Table V of the main text, the initial learning rate is set as $2\times10^{-6}$, the batch size is 1, the image size of the Transformer backbone is $128\times 128$, the patch size is 2, the hidden size is 768, the depth is 12, and the number of head is 12. The first stage model and the conditioning stage model share a same VQGAN model, with the parameters of the model using pretrained vq-f4 from LDM. The training epoch of the DiT model is 99.

For the CycleGAN model in Table V of the main text, the initial learning rate is set as $5\times10^{-5}$, the batch size is 2, the architecture of the generator is resnet with 9 blocks. The default configuration of CycleGAN is used for other parameter settings. The training epoch of the CycleGAN model is 200.

For the HRIGNet in Table V of the main text, the training epoch is 98.

\subsection{Ablation Study}
\label{append:experiments_ablation}

\begin{table}[t]\centering
    \caption{Ablation study on the guiding model of HRIGNet.}
    \resizebox*{0.48\textwidth}{!}{
        \begin{tabular}{cccccc}
            \hline
            \makecell[c]{ guiding \\ information } & resolution & FID$\downarrow$ & LPIPS$\downarrow$ & SSIM$\uparrow$ & PSNR$\uparrow$ \\ 
            \hline
            Rainy Image & 256$\times$256 & 150.851 & 0.505 & 0.389 & 13.692 \\ 
            Rain Layer & 256$\times$256 & 149.779 & 0.506 & 0.390 & 13.674 \\ 
            Rain Layer  & 512$\times$512 & \textbf{32.111} & \textbf{0.205} & \textbf{0.747} & \textbf{18.595} \\
            \hline
        \end{tabular}
    }
    \label{tab:table_guiding}
\end{table}

\begin{table}[t]\centering
    \caption{Ablation study on the backbone of HRIGNet.}
    \resizebox*{0.48\textwidth}{!}{
        \begin{tabular}{cccccc}
            \hline
            backbone & resolution & FID$\downarrow$ & LPIPS$\downarrow$ & SSIM$\uparrow$ & PSNR$\uparrow$ \\ 
            \hline
            Tramsformer & 512$\times$512 & 288.696 & 0.612 & 0.515 & 15.809  \\ 
            UNet & 512$\times$512 & \textbf{32.111} & \textbf{0.205} & \textbf{0.747} & \textbf{18.595} \\ 
            \hline
        \end{tabular}
    }
    \label{tab:table_backbone}
\end{table}

\textbf{Guiding ablation.} We conduct an ablation study to evaluate the effect of using guiding models with different guiding information and resolutions. Specifically, we compare the performance of using rainy image and rain layer as guiding information, and resolution of $256\times 256$ and $512\times 512$ in HRIGNet. As shown in Table \ref{tab:table_guiding}, the model using rain layer of size $512\times 512$ as guidance achieves the best result in all metrics.

For the HRIGNet model with rainy image of size $256\times 256$ as guidance in Table \ref{tab:table_guiding}, the initial learning rate is set as $2\times10^{-6}$, the batch size is 1, the image size of the UNet backbone is $128\times 128$, and the model channels are 224. The first stage model and the conditioning stage model share a same VQGAN model, with the parameters of the model using pretrained vq-f4 from LDM. For the guiding diffusion model, the image size of the UNet backbone is $64\times 64$, and the model channels are 224. The training epoch of the guiding diffusion model is 99, and the training epoch of the HRIGNet model is 88.

For the HRIGNet model with rain layer of size $256\times 256$ as guidance in Table \ref{tab:table_guiding}, the initial learning rate is set as $2\times10^{-6}$, the batch size is 1, the image size of the UNet backbone is $128\times 128$, and the model channels are 224. The first stage model and the conditioning stage model share a same VQGAN model, with the parameters of the model using pretrained vq-f4 from LDM. For the guiding diffusion model, the image size of the UNet backbone is $64\times 64$, and the model channels are 224. The training epoch of the guiding diffusion model is 96, and the training epoch of the HRIGNet model is 88.

For the guiding diffusion model of the HRIGNet model with rain layer of size $512\times 512$ as guidance in Table \ref{tab:table_guiding}, the image size of the UNet backbone is $128\times 128$, and the model channels are 224. The training epoch of the guiding diffusion model is 99.

\textbf{Backbone ablation.} We also conduct an ablation study to investigate the effect of using different backbones of the diffusion model on HRIGNet. Specifically, we compare two popular backbone architectures, UNet \cite{rombach2022high} and Transformer \cite{peebles2023scalable}. As shown in Table \ref{tab:table_backbone}, the results indicate that HRIGNet using UNet backbone outperforms the one using Transformer backbone in all metrics. According to DiT \cite{peebles2023scalable}, the scaling properties of the Transformer can extend to diffusion models with Transformer backbones. However, models with Transformer backbones tend to underperform when the model size is inadequate due to their inherent scaling properties. Furthermore, in the experiments, we also find that models with Transformer backbones have a slower convergence speed. Therefore, we finally adopt the UNet backbone in our model.

For the HRIGNet model with Transformer backbone in Table \ref{tab:table_backbone}, the initial learning rate is set as $2\times10^{-6}$, the batch size is 1. For the Transformer backbone, the image size is $128\times 128$, the patch size is 8, the hidden size is 384, the depth is 12, and the number of head is 12. The first stage model and the conditioning stage model share a same VQGAN model, with the parameters of the model using pretrained vq-f4 from LDM. The training epoch of the HRIGNet model is 75.

\bibliographystyle{IEEEtran}
\bibliography{refs}

\begin{thebibliography}{10}
\providecommand{\url}[1]{#1}
\csname url@samestyle\endcsname
\providecommand{\newblock}{\relax}
\providecommand{\bibinfo}[2]{#2}
\providecommand{\BIBentrySTDinterwordspacing}{\spaceskip=0pt\relax}
\providecommand{\BIBentryALTinterwordstretchfactor}{4}
\providecommand{\BIBentryALTinterwordspacing}{\spaceskip=\fontdimen2\font plus
\BIBentryALTinterwordstretchfactor\fontdimen3\font minus \fontdimen4\font\relax}
\providecommand{\BIBforeignlanguage}[2]{{%
\expandafter\ifx\csname l@#1\endcsname\relax
\typeout{** WARNING: IEEEtran.bst: No hyphenation pattern has been}%
\typeout{** loaded for the language `#1'. Using the pattern for}%
\typeout{** the default language instead.}%
\else
\language=\csname l@#1\endcsname
\fi
#2}}
\providecommand{\BIBdecl}{\relax}
\BIBdecl

\bibitem{chen2022vision}
Z.~Chen, Y.~Duan, W.~Wang, J.~He, T.~Lu, J.~Dai, and Y.~Qiao, ``Vision transformer adapter for dense predictions,'' \emph{arXiv preprint arXiv:2205.08534}, 2022.

\bibitem{wang2023internimage}
W.~Wang, J.~Dai, Z.~Chen, Z.~Huang, Z.~Li, X.~Zhu, X.~Hu, T.~Lu, L.~Lu, H.~Li \emph{et~al.}, ``Internimage: Exploring large-scale vision foundation models with deformable convolutions,'' in \emph{Proceedings of the IEEE/CVF conference on computer vision and pattern recognition}, 2023, pp. 14\,408--14\,419.

\bibitem{jain2023semask}
J.~Jain, A.~Singh, N.~Orlov, Z.~Huang, J.~Li, S.~Walton, and H.~Shi, ``Semask: Semantically masked transformers for semantic segmentation,'' in \emph{Proceedings of the IEEE/CVF international conference on computer vision}, 2023, pp. 752--761.

\bibitem{hassani2023neighborhood}
A.~Hassani, S.~Walton, J.~Li, S.~Li, and H.~Shi, ``Neighborhood attention transformer,'' in \emph{Proceedings of the IEEE/CVF conference on computer vision and pattern recognition}, 2023, pp. 6185--6194.

\bibitem{sakaridis2021acdc}
C.~Sakaridis, D.~Dai, and L.~Van~Gool, ``Acdc: The adverse conditions dataset with correspondences for semantic driving scene understanding,'' in \emph{Proceedings of the IEEE/CVF international conference on computer vision}, 2021, pp. 10\,765--10\,775.

\bibitem{sun2022shift}
T.~Sun, M.~Segu, J.~Postels, Y.~Wang, L.~Van~Gool, B.~Schiele, F.~Tombari, and F.~Yu, ``Shift: a synthetic driving dataset for continuous multi-task domain adaptation,'' in \emph{Proceedings of the IEEE/CVF Conference on Computer Vision and Pattern Recognition}, 2022, pp. 21\,371--21\,382.

\bibitem{westra2014future}
S.~Westra, H.~J. Fowler, J.~P. Evans, L.~V. Alexander, P.~Berg, F.~Johnson, E.~J. Kendon, G.~Lenderink, and N.~Roberts, ``Future changes to the intensity and frequency of short-duration extreme rainfall,'' \emph{Reviews of Geophysics}, vol.~52, no.~3, pp. 522--555, 2014.

\bibitem{garg2006photorealistic}
K.~Garg and S.~K. Nayar, ``Photorealistic rendering of rain streaks,'' \emph{ACM Transactions on Graphics (TOG)}, vol.~25, no.~3, pp. 996--1002, 2006.

\bibitem{halder2019physics}
S.~S. Halder, J.-F. Lalonde, and R.~d. Charette, ``Physics-based rendering for improving robustness to rain,'' in \emph{Proceedings of the IEEE/CVF International Conference on Computer Vision}, 2019, pp. 10\,203--10\,212.

\bibitem{wang2021rain}
H.~Wang, Z.~Yue, Q.~Xie, Q.~Zhao, Y.~Zheng, and D.~Meng, ``From rain generation to rain removal,'' in \emph{Proceedings of the IEEE/CVF Conference on Computer Vision and Pattern Recognition}, 2021, pp. 14\,791--14\,801.

\bibitem{dosovitskiy2017carla}
A.~Dosovitskiy, G.~Ros, F.~Codevilla, A.~Lopez, and V.~Koltun, ``Carla: An open urban driving simulator,'' in \emph{Conference on robot learning}.\hskip 1em plus 0.5em minus 0.4em\relax PMLR, 2017, pp. 1--16.

\bibitem{shah2018airsim}
S.~Shah, D.~Dey, C.~Lovett, and A.~Kapoor, ``Airsim: High-fidelity visual and physical simulation for autonomous vehicles,'' in \emph{Field and Service Robotics: Results of the 11th International Conference}.\hskip 1em plus 0.5em minus 0.4em\relax Springer, 2018, pp. 621--635.

\bibitem{yang2017deep}
W.~Yang, R.~T. Tan, J.~Feng, J.~Liu, Z.~Guo, and S.~Yan, ``Deep joint rain detection and removal from a single image,'' in \emph{Proceedings of the IEEE Conference on Computer Vision and Pattern Recognition}, 2017, pp. 1357--1366.

\bibitem{hu2019depth}
X.~Hu, C.-W. Fu, L.~Zhu, and P.-A. Heng, ``Depth-attentional features for single-image rain removal,'' in \emph{Proceedings of the IEEE/CVF Conference on Computer Vision and Pattern Recognition}, 2019, pp. 8022--8031.

\bibitem{jiang2020multi}
K.~Jiang, Z.~Wang, P.~Yi, C.~Chen, B.~Huang, Y.~Luo, J.~Ma, and J.~Jiang, ``Multi-scale progressive fusion network for single image deraining,'' in \emph{Proceedings of the IEEE/CVF Conference on Computer Vision and Pattern Recognition}, 2020, pp. 8346--8355.

\bibitem{ni2021controlling}
S.~Ni, X.~Cao, T.~Yue, and X.~Hu, ``Controlling the rain: From removal to rendering,'' in \emph{Proceedings of the IEEE/CVF Conference on Computer Vision and Pattern Recognition}, 2021, pp. 6328--6337.

\bibitem{ye2021closing}
Y.~Ye, Y.~Chang, H.~Zhou, and L.~Yan, ``Closing the loop: Joint rain generation and removal via disentangled image translation,'' in \emph{Proceedings of the IEEE/CVF Conference on Computer Vision and Pattern Recognition}, 2021, pp. 2053--2062.

\bibitem{zhou2024controllable}
K.~Zhou, S.~Zhao, and H.~Deng, ``Controllable rain image generation: Balance between diversity and controllability,'' in \emph{International Conference on Intelligent Computing}.\hskip 1em plus 0.5em minus 0.4em\relax Springer, 2024, pp. 86--97.

\bibitem{li2019single}
S.~Li, I.~B. Araujo, W.~Ren, Z.~Wang, E.~K. Tokuda, R.~H. Junior, R.~Cesar-Junior, J.~Zhang, X.~Guo, and X.~Cao, ``Single image deraining: A comprehensive benchmark analysis,'' in \emph{Proceedings of the IEEE/CVF Conference on Computer Vision and Pattern Recognition}, 2019, pp. 3838--3847.

\bibitem{quan2021removing}
R.~Quan, X.~Yu, Y.~Liang, and Y.~Yang, ``Removing raindrops and rain streaks in one go,'' in \emph{Proceedings of the IEEE/CVF Conference on Computer Vision and Pattern Recognition}, 2021, pp. 9147--9156.

\bibitem{neuhold2017mapillary}
G.~Neuhold, T.~Ollmann, S.~Rota~Bulo, and P.~Kontschieder, ``The mapillary vistas dataset for semantic understanding of street scenes,'' in \emph{Proceedings of the IEEE international conference on computer vision}, 2017, pp. 4990--4999.

\bibitem{caesar2020nuscenes}
H.~Caesar, V.~Bankiti, A.~H. Lang, S.~Vora, V.~E. Liong, Q.~Xu, A.~Krishnan, Y.~Pan, G.~Baldan, and O.~Beijbom, ``nuscenes: A multimodal dataset for autonomous driving,'' in \emph{Proceedings of the IEEE/CVF conference on computer vision and pattern recognition}, 2020, pp. 11\,621--11\,631.

\bibitem{cordts2016cityscapes}
M.~Cordts, M.~Omran, S.~Ramos, T.~Rehfeld, M.~Enzweiler, R.~Benenson, U.~Franke, S.~Roth, and B.~Schiele, ``The cityscapes dataset for semantic urban scene understanding,'' in \emph{Proceedings of the IEEE conference on computer vision and pattern recognition}, 2016, pp. 3213--3223.

\bibitem{goodfellow2020generative}
I.~Goodfellow, J.~Pouget-Abadie, M.~Mirza, B.~Xu, D.~Warde-Farley, S.~Ozair, A.~Courville, and Y.~Bengio, ``Generative adversarial networks,'' \emph{Communications of the ACM}, vol.~63, no.~11, pp. 139--144, 2020.

\bibitem{zhu2017unpaired}
J.-Y. Zhu, T.~Park, P.~Isola, and A.~A. Efros, ``Unpaired image-to-image translation using cycle-consistent adversarial networks,'' in \emph{Proceedings of the IEEE International Conference on Computer Vision}, 2017, pp. 2223--2232.

\bibitem{kingma2013auto}
D.~P. Kingma and M.~Welling, ``Auto-encoding variational bayes,'' \emph{arXiv preprint arXiv:1312.6114}, 2013.

\bibitem{jaipuria2020deflating}
N.~Jaipuria, X.~Zhang, R.~Bhasin, M.~Arafa, P.~Chakravarty, S.~Shrivastava, S.~Manglani, and V.~N. Murali, ``Deflating dataset bias using synthetic data augmentation,'' in \emph{Proceedings of the IEEE/CVF Conference on Computer Vision and Pattern Recognition Workshops}, 2020, pp. 772--773.

\bibitem{li2025improving}
H.~Li, H.~K. Chu, and Y.~Sun, ``Improving rgb-thermal semantic scene understanding with synthetic data augmentation for autonomous driving,'' \emph{IEEE Robotics and Automation Letters}, 2025.

\bibitem{abu2018augmented}
H.~Abu~Alhaija, S.~K. Mustikovela, L.~Mescheder, A.~Geiger, and C.~Rother, ``Augmented reality meets computer vision: Efficient data generation for urban driving scenes,'' \emph{International Journal of Computer Vision}, vol. 126, pp. 961--972, 2018.

\bibitem{zheng2023robust}
Z.~Zheng, Y.~Cheng, Z.~Xin, Z.~Yu, and B.~Zheng, ``Robust perception under adverse conditions for autonomous driving based on data augmentation,'' \emph{IEEE Transactions on Intelligent Transportation Systems}, vol.~24, no.~12, pp. 13\,916--13\,929, 2023.

\bibitem{sanders2016introduction}
A.~Sanders, \emph{An introduction to Unreal engine 4}.\hskip 1em plus 0.5em minus 0.4em\relax AK Peters/CRC Press, 2016.

\bibitem{pharr2016physically}
M.~Pharr, W.~Jakob, and G.~Humphreys, \emph{Physically based rendering: From theory to implementation}.\hskip 1em plus 0.5em minus 0.4em\relax Morgan Kaufmann, 2016.

\bibitem{blender2018}
\BIBentryALTinterwordspacing
B.~O. Community, \emph{Blender - a 3D modelling and rendering package}, Blender Foundation, Stichting Blender Foundation, Amsterdam, 2018. [Online]. Available: \url{http://www.blender.org}
\BIBentrySTDinterwordspacing

\bibitem{duhanyan2011below}
N.~Duhanyan and Y.~Roustan, ``Below-cloud scavenging by rain of atmospheric gases and particulates,'' \emph{Atmospheric Environment}, vol.~45, no.~39, pp. 7201--7217, 2011.

\bibitem{best1950size}
A.~Best, ``The size distribution of raindrops,'' \emph{Quarterly Journal of the Royal Meteorological Society}, vol.~76, no. 327, pp. 16--36, 1950.

\bibitem{kessler1969distribution}
E.~Kessler, ``On the distribution and continuity of water substance in atmospheric circulations,'' in \emph{On the Distribution and Continuity of Water Substance in Atmospheric Circulations}.\hskip 1em plus 0.5em minus 0.4em\relax Springer, 1969, pp. 1--84.

\bibitem{rombach2022high}
R.~Rombach, A.~Blattmann, D.~Lorenz, P.~Esser, and B.~Ommer, ``High-resolution image synthesis with latent diffusion models,'' in \emph{Proceedings of the IEEE/CVF Conference on Computer Vision and Pattern Recognition}, 2022, pp. 10\,684--10\,695.

\bibitem{esser2021taming}
P.~Esser, R.~Rombach, and B.~Ommer, ``Taming transformers for high-resolution image synthesis,'' in \emph{Proceedings of the IEEE/CVF Conference on Computer Vision and Pattern Recognition}, 2021, pp. 12\,873--12\,883.

\bibitem{ronneberger2015u}
O.~Ronneberger, P.~Fischer, and T.~Brox, ``U-net: Convolutional networks for biomedical image segmentation,'' in \emph{Medical Image Computing and Computer-Assisted Intervention--MICCAI 2015: 18th International Conference, Munich, Germany, October 5-9, 2015, Proceedings, Part III 18}.\hskip 1em plus 0.5em minus 0.4em\relax Springer, 2015, pp. 234--241.

\bibitem{peebles2023scalable}
W.~Peebles and S.~Xie, ``Scalable diffusion models with transformers,'' in \emph{Proceedings of the IEEE/CVF International Conference on Computer Vision}, 2023, pp. 4195--4205.

\bibitem{vaswani2017attention}
A.~Vaswani, N.~Shazeer, N.~Parmar, J.~Uszkoreit, L.~Jones, A.~N. Gomez, {\L}.~Kaiser, and I.~Polosukhin, ``Attention is all you need,'' \emph{Advances in Neural Information Processing Systems}, vol.~30, 2017.

\bibitem{lee2024unreal}
N.~Lee, ``Unreal engine, a 3d game engine,'' in \emph{Encyclopedia of Computer Graphics and Games}.\hskip 1em plus 0.5em minus 0.4em\relax Springer, 2024, pp. 1944--1947.

\bibitem{zhang2023adding}
L.~Zhang, A.~Rao, and M.~Agrawala, ``Adding conditional control to text-to-image diffusion models,'' in \emph{Proceedings of the IEEE/CVF international conference on computer vision}, 2023, pp. 3836--3847.

\bibitem{heusel2017gans}
M.~Heusel, H.~Ramsauer, T.~Unterthiner, B.~Nessler, and S.~Hochreiter, ``Gans trained by a two time-scale update rule converge to a local nash equilibrium,'' \emph{Advances in Neural Information Processing Systems}, vol.~30, 2017.

\bibitem{zhang2018unreasonable}
R.~Zhang, P.~Isola, A.~A. Efros, E.~Shechtman, and O.~Wang, ``The unreasonable effectiveness of deep features as a perceptual metric,'' in \emph{Proceedings of the IEEE Conference on Computer Vision and Pattern Recognition}, 2018, pp. 586--595.

\bibitem{ren2019progressive}
D.~Ren, W.~Zuo, Q.~Hu, P.~Zhu, and D.~Meng, ``Progressive image deraining networks: A better and simpler baseline,'' in \emph{Proceedings of the IEEE/CVF Conference on Computer Vision and Pattern Recognition}, 2019, pp. 3937--3946.

\bibitem{gao2023mountain}
H.~Gao, J.~Yang, Y.~Zhang, N.~Wang, J.~Yang, and D.~Dang, ``A mountain-shaped single-stage network for accurate image restoration,'' \emph{arXiv preprint arXiv:2305.05146}, 2023.

\bibitem{cui2022selective}
Y.~Cui, Y.~Tao, Z.~Bing, W.~Ren, X.~Gao, X.~Cao, K.~Huang, and A.~Knoll, ``Selective frequency network for image restoration,'' in \emph{The Eleventh International Conference on Learning Representations}, 2022.

\bibitem{zamir2022restormer}
S.~W. Zamir, A.~Arora, S.~Khan, M.~Hayat, F.~S. Khan, and M.-H. Yang, ``Restormer: Efficient transformer for high-resolution image restoration,'' in \emph{Proceedings of the IEEE/CVF Conference on Computer Vision and Pattern Recognition}, 2022, pp. 5728--5739.

\bibitem{fu2017removing}
X.~Fu, J.~Huang, D.~Zeng, Y.~Huang, X.~Ding, and J.~Paisley, ``Removing rain from single images via a deep detail network,'' in \emph{Proceedings of the IEEE Conference on Computer Vision and Pattern Recognition}, 2017, pp. 3855--3863.

\bibitem{wang2019spatial}
T.~Wang, X.~Yang, K.~Xu, S.~Chen, Q.~Zhang, and R.~W. Lau, ``Spatial attentive single-image deraining with a high quality real rain dataset,'' in \emph{Proceedings of the IEEE/CVF Conference on Computer Vision and Pattern Recognition}, 2019, pp. 12\,270--12\,279.

\end{thebibliography}

\end{document}